%% file: main.tex
\documentclass[10pt,twocolumn,letterpaper,usenames,dvipsnames]{article}

\usepackage{iccv}
\usepackage{times}
\usepackage{epsfig}
\usepackage{graphicx}
\usepackage{float}
\usepackage{amsmath}
\usepackage{amssymb}
\usepackage{soul}
\usepackage{multicol}
\usepackage{blindtext}
\usepackage{here}
\usepackage{caption}
\usepackage{makecell}
\usepackage[normalem]{ulem}
\usepackage{multirow}
\usepackage{booktabs}
\usepackage{xcolor, colortbl}
\usepackage{xspace}
\usepackage{balance}
\usepackage{numprint}
\usepackage{sidecap}
\usepackage[numbers,sort,compress]{natbib}

\definecolor{Gray}{gray}{0.9}
\usepackage[pagebackref=true,breaklinks=true,letterpaper=true,colorlinks,bookmarks=false]{hyperref}
\iccvfinalcopy % *** Uncomment this line for the final
\def\iccvPaperID{1111} % *** Enter the CVPR Paper ID here

\ificcvfinal\fi

\begin{document}

%%%%%%%%% TITLE
\title{SPEC: Seeing People in the Wild with an Estimated Camera}

% \author{
% Anonymous Authors
% }
\author{%
  Muhammed Kocabas$^{1,2}$\quad \; Chun-Hao P. Huang$^1$\quad \; Joachim Tesch$^1$ \quad \; Lea M\"uller$^1$ \quad \; \\ Otmar Hilliges$^2$\quad \; Michael J. Black$^1$\\\
  \normalsize $^1$Max Planck Institute for Intelligent Systems, T\"{u}bingen, Germany \quad
  \normalsize $^2$ETH Zurich\\
  \normalsize \texttt{\{\href{mailto:mkocabas@tue.mpg.de}{mkocabas},\href{mailto:nathanasiou@tue.mpg.de}{chuang2},\href{mailto:joachim.tesch@tuebingen.mpg.de}{jtesch},\href{mailto:lea.mueller@tuebingen.mpg.de}{lea.mueller},\href{mailto:black@tue.mpg.de}{black}\}@tue.mpg.de} \quad 
  \texttt{\href{mailto:otmar.hilliges@inf.ethz.ch}{otmar.hilliges@inf.ethz.ch} }
}
\input{sections/10_macros}
% \ificcvfinal\pagestyle{empty}\fi

\twocolumn[{%
	\renewcommand\twocolumn[1][]{#1}%
	\maketitle
	\begin{center}
		\newcommand{\teaserwidth}{\textwidth}
		\vspace{-0.15in}
		\centerline{
			\includegraphics[width=\teaserwidth,clip]{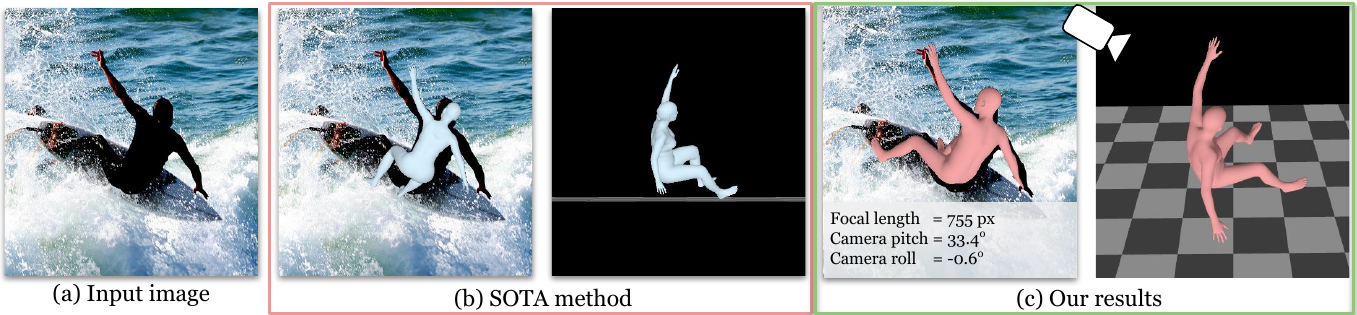}
		}
		\vspace{-1ex}
		\captionof{figure}{State-of-the-art methods for 3D human pose and shape estimation from images (such as HMR trained with EFT~\cite{joo2020eft} data) struggle with imagery containing perspective effects. (b) In part, this is due to the use of the standard weak perspective camera. (c) \methodname learns to estimate perspective camera parameters and uses these to regress more accurate 3D poses.}
		\vspace{-0.1in}
		\label{fig:teaser}
	\end{center}%
}]
\input{sections/0_abstract.tex}
\input{sections/1_introduction.tex}

\input{sections/2_related_work.tex}
\input{sections/3_methodology.tex}
\input{sections/4_experiments.tex}
\input{sections/5_conclusion.tex}
\input{sections/8_acknowledgements.tex}

{\small
\bibliographystyle{ieee_fullname}
\balance
\bibliography{main}
}
	\newpage
	%%%%%%%%% TITLE
	\appendix
	{\noindent\Large\textbf{Supplementary Material}}
	\newline
	\setcounter{page}{1}
	
	In this supplementary document, we provide more information that is not covered in the main text, ranging from technical details in the method, visual examples of the proposed datasets, more qualitative results, more ablation studies as well as more analysis and discussion.

    \input{sections/9_supmat}
\end{document}

%% file: sections/10_macros.tex
\newcommand{\mkocabas}[1]{\textcolor{cyan}{[MK: {#1}]}}
\newcommand{\eaksan}[1]{\textcolor{blue}{[EA: {#1}]}}
\newcommand{\phuang}[1]{\textcolor{red}{[PH: {#1}]}}
\newcommand{\otmar}[1]{\textcolor{orange}{[OH: {#1}]}}
\newcommand{\michael}[1]{\textcolor{magenta}{[MJB: {#1}]}}
\newcommand{\lea}[1]{\textcolor{olive}{[LM: {#1}]}}
\newcommand{\oh}[1]{\otmar{#1}}
% Math notations for consistency

\newcommand{\predTheta}{\hat{\mathbf{\Theta}}}
\newcommand{\predShape}{\hat{\mathbf{\beta}}}
\newcommand{\predPose}{\hat{\mathbf{\theta}}}
	
\newcommand{\gtTheta}{\mathbf{\Theta}}
\newcommand{\gtShape}{\mathbf{\beta}}
\newcommand{\gtPose}{\mathbf{\theta}}

\newcommand{\motionDisc}{\mathcal{D}_M}	
\newcommand{\generator}{\mathcal{G}}
\newcommand{\smpl}{\mathcal{M}}	

\newcommand{\focalx}{f_x}
\newcommand{\focaly}{f_y}	
\newcommand{\centerx}{o_x}
\newcommand{\centery}{o_y}
\newcommand{\campos}{C}
\newcommand{\camtransl}{t^c}
\newcommand{\camrot}{R^c}
\newcommand{\imgwidth}{w}
\newcommand{\imgheight}{h}
\newcommand{\cropwidth}{w_{bbox}}
\newcommand{\cropheight}{h_{bbox}}
\newcommand{\cropcenterx}{c_x}
\newcommand{\cropcentery}{c_y}
\newcommand{\predjoints}[1]{\hat{\mathcal{J}}_{\mathit{#1D}}}
\newcommand{\gtjoints}[1]{\mathcal{J}_{\mathit{#1D}}}

\newcommand{\bodytransl}{t^b}
\newcommand{\bodyori}{R^b}

\newcommand{\pampjpe}{PA-MPJPE\xspace}
\newcommand{\wmpjpe}{W-MPJPE\xspace}
\newcommand{\cmpjpe}{C-MPJPE\xspace}
\newcommand{\mpjpe}{MPJPE\xspace}
\newcommand{\wpve}{W-PVE\xspace}
\newcommand{\cpve}{C-PVE\xspace}
\newcommand{\pve}{PVE\xspace}

\newcommand{\iwcam}{IWP-cam\xspace}
\newcommand{\camcalib}{CamCalib\xspace}
\newcommand{\pitch}{\alpha}
\newcommand{\roll}{\phi}
\newcommand{\yaw}{\psi}
\newcommand{\vfov}{\upsilon}
\newcommand{\softltwo}{Softargmax-$\mathcal{L}_{2}$\xspace}
\newcommand{\softbiasedltwo}{Softargmax-biased-$\mathcal{L}_{2}$\xspace}
\newcommand{\ltwo}{$\mathcal{L}_2$\xspace}
\newcommand{\smplify}{SMPLify-X-cam\xspace}

\newcommand{\methodname}{SPEC\xspace} % Use this as to refer our method in case we change it in the future
% \DeclareMathOperator*{\argmin}{arg\,min}
%% useful sjpothands
\newcommand{\figref}[1]{Fig.~\ref{#1}}

%%%% DATASET NAMES %%%%
\newcommand{\mpi}{\texttt{MPI-INF-3DHP}\xspace}
\newcommand{\mpii}{\texttt{MPII}\xspace}
\newcommand{\lspet}{\texttt{LSPET}\xspace}
\newcommand{\hthreesixm}{\texttt{Human3.6M}\xspace}
\newcommand{\threedpw}{\texttt{3DPW}\xspace}
\newcommand{\threedpwocc}{\texttt{3DPW-OCC}\xspace}
\newcommand{\coco}{\texttt{COCO}\xspace}
\newcommand{\cocoeft}{\texttt{COCO-EFT}\xspace}
\newcommand{\ooh}{\texttt{3DOH}\xspace}

\newcommand{\agoracam}{\methodname-SYN\xspace}
\newcommand{\mtpcam}{\methodname-MTP\xspace}

\newcommand{\smplifyxc}{SMPLify-XC\xspace}

\newcommand{\supmat}{Sup.~Mat.\xspace}

\newcommand{\real}{\mathbb{R}}

% \cvprfinalcopy % *** Uncomment this line for the final submission
\renewcommand{\etal}{et al.\xspace}
\renewcommand{\ie}{i.e.\xspace}
\renewcommand{\eg}{e.g.\xspace}

%% file: sections/0_abstract.tex
%%%%%%%%% ABSTRACT
\begin{abstract}
Due to the lack of camera parameter information for in-the-wild images, existing 3D human pose and shape (HPS) estimation methods make several simplifying assumptions: weak-perspective projection, large constant focal length, and zero camera rotation.
These  assumptions often do not hold and we show, 
quantitatively and qualitatively, 
that they cause errors in the reconstructed 3D shape and pose. 
To address this, we introduce \methodname, the first in-the-wild 3D HPS method that estimates the perspective camera from a single image and employs this to reconstruct 3D human bodies more accurately. %regress 3D human bodies.
First, we train a neural network to estimate the field of view, camera pitch, and roll given an input image.
We employ novel losses that improve the calibration accuracy over previous work.
We then train a novel network that concatenates the camera calibration to the image features and uses these together to regress 3D body shape and pose.
\methodname is more accurate than the prior art on the standard benchmark (3DPW) as well as two new datasets with more challenging camera views and varying focal lengths.
Specifically, we create a new photorealistic synthetic dataset (\agoracam) with ground truth 3D bodies and a novel in-the-wild dataset (\mtpcam) with calibration and high-quality reference bodies.
%
%Both qualitative and quantitative analysis confirm that knowing camera parameters during inference regresses better human bodies.
Code and datasets are available for research purposes at {\small \url{https://spec.is.tue.mpg.de/}}.
%\oh{TODO: preview of the most important results}
\end{abstract}

%% file: sections/1_introduction.tex
\vspace{-2ex}
\section{Introduction}
\label{introduction}
Estimating 3D human pose and shape (HPS) from a single RGB image is a core challenge in computer vision and has many  applications in robotics, computer graphics, and AR/VR. 
Reconstructing a high dimensional 3D structure from 2D observations is ill-posed by nature. 
To overcome this, much attention has been given to structured prediction \cite{deepcap,huang2017pami,vlasic2008articulated} and incorporating shape and pose priors \cite{SMPL-X:2019,Xu_2020_CVPR} to guide estimation.
Weakly-supervised training of HPS regressors leverages 2D-pose datasets \cite{mpii,lspet,coco} and requires various forms of regularization \cite{kanazawa_hmr,SPIN:ICCV:2019,xu2019denserac}.
Data for full 3D supervision often relies on 
controlled lab settings \cite{ionescu_h36m,Sigal:IJCV:10b},  synthetic images \cite{varol17_surreal}, or, more recently, in-the-wild capture of reference data \cite{Mehta2018XNectRM,vonMarcard2018_3dpw}.

Despite rapid progress, we observe that most state-of-the-art (SOTA) methods \cite{bogo_smplify,expose2020eccv,guler_2019_CVPR,jiang2020mpshape,joo2020eft,kanazawa_hmr,kanazawa_temporal_hmr,kocabas2019vibe,kolotouros2019cmr,SPIN:ICCV:2019,SMPL-X:2019,Rockwell2020,song2020human,CenterHMR,zanfir2020weakly} make several simplifying assumptions about the image formation process itself.
First, they all apply a weak perspective or orthographic projection assumption; resulting in a simplified camera model with only three parameters which capture the camera translation relative to the body.
Moreover, some \cite{kocabas2019vibe,SPIN:ICCV:2019,SMPL-X:2019} set the focal length to a predefined large constant for every input image.
Finally, they all assume zero camera rotation, which entangles body rotation and camera rotation, making it extremely hard to correctly estimate the body orientation in 3D.
These assumptions are valid for images where bodies are roughly perpendicular to the principal axis and are located far away from the camera. %\eg Fig.~\ref{fig:naive_cam} (c)\textcolor{red}{2-c is missing}. 
However, in most real world images of people, perspective effects are clearly evident, \eg~foreshortening in selfies. 
Ignoring perspective projection leads to errors in pose, shape, and global orientation (see Fig.~\ref{fig:teaser}). 

To overcome these limitations in existing methods, we present \emph{\methodname (Seeing People in the wild with Estimated Cameras)}, the first 3D human pose and shape estimation framework that leverages cues present in the image to extract perspective camera information and exploits this to better reconstruct 3D human bodies from images in the wild.
\methodname consists of two parts: camera calibration and body reconstruction. We make contributions to each.

One might hope that  embedded EXIF information would be sufficient to address this problem.
However, many images lack EXIF information, some applications strip this off, and even if present, converting the stored focal length in millimeters to pixels requires knowing specifics of the image sensor.
Given the huge variety of cameras on the market, exploiting this is a non-trivial task.
Furthermore, this does not give information about the camera rotation. %\oh{This seems like a secondary argument and sounds defensive. I would consider moving or dropping entirely. Could be in related work, discussion or supp. mat.}

Instead, we estimate the camera directly from the RGB image.
Recent work \cite{Hold-Geoffroy_2018_CVPR,Kendall_2015_ICCV,workman2016horizon,zhu2020single} casts this ill-posed regression problem as a classification task.
However, training such methods with their losses, \eg~cross-entropy and KL-divergence, 
%\oh{KL-D is not per-se categorical},
ignores the natural notion of distance or ordering of the original target space.
To address this, we propose a new loss, \softltwo, to preserve distance during loss calculation. Moreover, we observe that HPS accuracy is quite sensitive to underestimation of focal length and less sensitive to overestimates as also noted by ~\cite{kissosECCVW2020,yu2020pcls}. Therefore, we modify \softltwo to be asymmetric such that less penalty is applied when the focal length is overestimated. These novel losses help us to train a better regressor for direct camera calibration, which we term \emph{\camcalib}.

We integrate the regressed camera parameters into two 3D-body-reconstruction paradigms: (1) an optimization-based approach, SMPLify-X \cite{SMPL-X:2019}, and (2) a regression-based one similar to HMR or SPIN \cite{kanazawa_hmr,SPIN:ICCV:2019}.
Since SMPLify-X estimates a 3D body my minimizing the difference between projected 3D joints and observed 2D joints, improving the the projective geometry improves the estimated body.

In the case of direct HPS regression from pixels, the estimated camera is employed in two ways: (1) in the reprojection loss similar to the one in SMPLify-X and (2) as {\em conditioning} for the network by appending the camera parameters to the CNN image features.
This second contribution is a key novelty of \methodname, which enables us to disentangle camera and body orientation. SOTA methods~\cite{PARE_2021, SPIN:ICCV:2019, kocabas2019vibe, CenterHMR}, cannot do this because the body is estimated in camera space, entangling body orientation and camera rotation.

Training such a body regressor requires in-the-wild images annotated with both 3D human bodies and the camera parameters.
Since existing 3D human body datasets \cite{ionescu_h36m,mpiiinf3dhp_mono-2017,vonMarcard2018_3dpw} contain little variation in camera parameters, we create two new datasets with rich camera variety. First, we create a photorealistic synthetic dataset which has accurate ground-truth human and camera annotations (\agoracam) using ideas from \cite{patel2021agora}. This dataset is used both for testing and training. Second, we collect a crowdsourced dataset following the Mimic-The-Pose framework \cite{Mueller:CVPR:21} (\mtpcam). 
We ask web participants to calibrate their camera and take videos from different angles while mimicking a predefined pose. Then, we obtain pseudo ground-truth labels by fitting the SMPL model to the provided videos while exploiting the predefined pose as a prior.
Through extensive experiments and analysis using these new datasets, alongside an existing in-the-wild dataset (3DPW \cite{vonMarcard2018_3dpw}), we show that going beyond the weak-perspective/orthographic assumption improves human pose and shape estimation results.

In summary, our contributions are:
%\begin{itemize}
    (1) We propose a single-view, camera-aware, 3D human body estimation framework that estimates perspective camera parameters from in-the-wild images directly and reconstructs the 3D body without relying on weak-perspective assumptions or offline calibration. 
    %\mkocabas{Shall we remove "first`` in case we missed a related work, or say "to the best of our knowledge``}
    (2) We train a neural network to regress the perspective camera parameters given one RGB image, using two novel losses: \softltwo and the asymmetric variant to improve the calibration accuracy. 
    (3) Using the estimated camera parameters  helps to reconstruct a better 3D body with the optimization-based SMPLify-X algorithm.
    (4) Conditioning on camera information helps a direct regression approach based on HMR \cite{kanazawa_hmr} learn to regress better poses.
    (5) We present two different datasets with ground-truth camera and human body parameters: (i) a photorealistic synthetic dataset, \agoracam, and (ii) a crowdsourced dataset, \mtpcam. % \textcolor{red}{Update.} %To obtain ground-truth data for our methods, 

%% file: sections/2_related_work.tex
\section{Related Work}
\label{related_work}
We review work that captures/reconstructs 3D humans under calibrated-camera settings, then focus on our goal: in-the-wild 3D human body reconstruction from monocular RGB. 
We discuss how prior art simplifies the camera model to make the problem tractable and also discuss relevant methods of camera calibration from one RGB image.

\noindent\textbf{3D HPS Estimation with Calibrated Cameras}.
To capture human motions in 3D, early work exploits calibrated and synchronized multi-camera settings.
They can be largely classified to ``bottom up'' approaches \cite{belagiannis20143d,burenius20133d,dong2019fast,grauman2003inferring,sigal2008combined,zhang20204d} 
that assemble 3D body poses from multi-view image evidence (keypoints, silhouettes),
and ``top down'' approaches \cite{Balan:CVPR:2007,EEJTP15,gall2009motion,huang2017pami,totalcapture,vlasic2008articulated} 
that deform a pre-defined 3D human template according to detected image features in each view.
Powered by CNNs, learning-based methods \cite{epipolartransformers2020cvpr,iskakov2019learnable,qiu2019crossview,remelli2020lightweight,to2020voxelpose} gain robustness by training keypoint detection and multi-view pose reconstruction end-to-end. 

Some monocular approaches \cite{mpiiinf3dhp_mono-2017,Mehta2018XNectRM,Pavlakos_2017_CVPR,shen2020multi} are trained with direct supervision from multi-view data, while others \cite{deepcap,iqbal2020cvpr,kocabas2019epipolar, rhodin2018learning} enforce multi-view consistency as weak, or self, supervision. 
In either way, known intrinsic or extrinsic parameters are always assumed.
Yu \etal~\cite{yu2020pcls} propose perspective crop layers, which crop the image around a person according to camera parameters and image location, effectively removing some effect of camera geometry.

All these methods require offline calibration and have a risk of overfitting to the cameras used in training and are typically limited to controlled settings.

\noindent\textbf{Single-view HPS Estimation with Unknown Cameras}.
In early work, Liebowitz and Carlsson \cite{Liebowitz:2001} use the repetitive structure of a moving person as a cue for camera calibration.
Since then, numerous methods reconstruct 3D human bodies given single-view images or videos in uncontrolled settings.
Closely related to structure-from-motion and bundle adjustment, \cite{arnab_kineticspose,dong2020motion,leroy2020smply,xiang2020monoclothcap} take videos as input 
and jointly estimate cameras and reconstruct human bodies;
\cite{hasler2009markerless, liu20204d} further ground the bodies in 3D scenes. 

We focus on the more general scenario in which the input is a single image. 
SOTA methods use parametric body models \cite{totalcapture,looper_smpl,SMPL-X:2019,xu2020ghum} 
and estimate the parameters either by fitting to detected image features \cite{bogo_smplify,SMPL-X:2019,xiang2019monocular} or by regressing directly from pixels with deep neural networks \cite{expose2020eccv,guler_2019_CVPR,jiang2020mpshape,joo2020eft,kanazawa_hmr,SPIN:ICCV:2019,Rockwell2020,rong2020frankmocap,song2020human,xu2019denserac,CenterHMR,zanfir2020weakly}.
All these approaches, including non-parametric approaches \cite{kolotouros2019cmr, saito2019pifu, saito2020pifuhd, Zeng_2020_CVPR}, assume weak perspective/orthographic projection or  pre-define the focal length as a large constant for all images. 
Additionally, they all assume zero camera rotation, which entangles body rotation and camera rotation.
As a result, these camera models have only three parameters, capturing the camera translation relative to the body.

Kissos \etal~\cite{kissosECCVW2020} identify this problem and show that replacing focal length with a constant closer to ground truth, \ie $f=5000 \rightarrow 2200$, improves results.
Wang et al.~\cite{wang2020viewpoint} demonstrate that jointly estimating camera viewpoints and 3D human poses improves cross-dataset generalization. 
To show this, they train a 3D pose estimation model on available 3D human pose datasets in a supervised way. 
However, these datasets are limited in terms of camera viewpoint and focal length diversity, background, and number of subjects.

In contrast to the above methods, \methodname generalizes to in-the-wild settings, varied camera intrinsics and viewpoints.

\noindent\textbf{Single-image Camera Calibration}.
Recent work \cite{Hold-Geoffroy_2018_CVPR,Kendall_2015_ICCV,workman2016horizon,zhu2020single} directly estimates camera parameters from a single image.
Zhu et al.~\cite{zhu2020single} also recover the height of some scene objects, \eg~people and cars, together with the camera geometry. 
They estimate 2D human poses but not 3D bodies.
To estimate camera rotations and fields of view, these methods train a neural network to leverage geometric cues in the image without calibration boards or body keypoints.
They discretize the continuous space of rotation into bins, casting the problem as a classification task and applying cross-entropy \cite{workman2016horizon} or KL-divergence \cite{Hold-Geoffroy_2018_CVPR,zhu2020single} losses. 
These losses unfortunately ignore the ordering in target spaces.
We devise new losses to retain the concept of distance in the original space, leading to better estimated cameras.

%% file: sections/3_methodology.tex
\begin{figure}[t]
    \centerline{
    \includegraphics[width=0.44\textwidth]{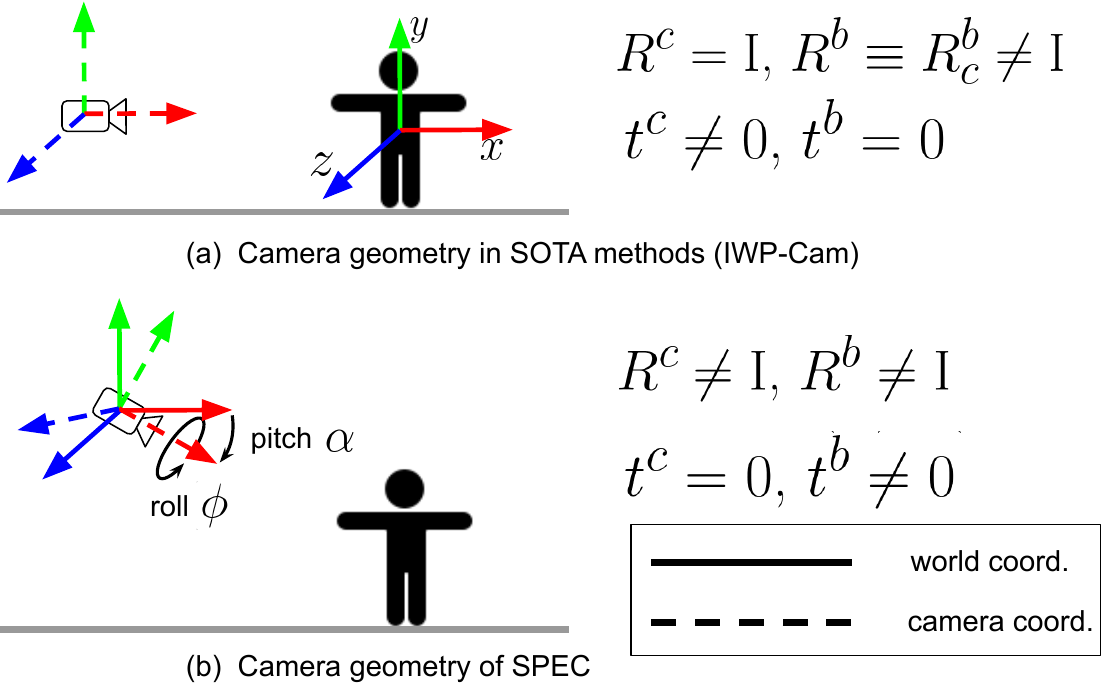}}
    \vspace{-0.1in}
    \caption{Illustration of \iwcam and \methodname. $\camrot$ and $\camtransl$ are camera rotation and translation. $\bodyori$ and $\bodytransl$ are body orientation and translation. All are defined in world coordinate.}
    \label{fig:naive_cam}
    \vspace{-2ex}
\end{figure}{}

\section{Method}
\label{methods}

\subsection{Preliminaries}
\label{recap}
A pinhole camera maps a 3D point $X \in \mathbb{R}^3$ to an image pixel $x \in \mathbb{R}^2$ through $x = K(\camrot X + \camtransl)$, where $K \in \mathbb{R}^{3\times 3}$ is the intrinsic matrix storing focal length $\focalx$, $\focaly$ and the principal point $(\centerx, \centery)$. 
We follow  previous work and omit skew, radial and tangential distortion. 
Extrinsic parameters are $\camrot \in SO(3)$ and $\camtransl = (t_x^c, t_y^c, t_z^c) \in \mathbb{R}^3$, representing camera rotation and translation in the world coordinate frame, respectively. %The camera location is $\campos=-\camrot ^ \top \camtransl$.

We estimate a parameteric human body model, SMPL \cite{looper_smpl,SMPL-X:2019}, that deforms a predefined human surface $\smpl$ according to body pose $\theta$ and shape $\beta$. When both body translation $\bodytransl$ and body orientation $\bodyori$ are zero, the mesh is located near the origin of world coordinates, facing the $z^{+}$ direction and $y^{+}$ is the up-vector, as visualized in Fig.~\ref{fig:naive_cam}(a).

Existing approaches \cite{bogo_smplify,expose2020eccv,guler_2019_CVPR,jiang2020mpshape,kanazawa_hmr,kocabas2019vibe,SMPL-X:2019,song2020human,CenterHMR} assume zero camera rotation, $\camrot = \text{I}$, and estimate the camera translation $\camtransl$ in two ways: 
(1) by fitting the body joint coordinates to 2D keypoints \cite{bogo_smplify,SMPL-X:2019,zanfir2020weakly} or 
(2) by predicting weak perspective camera parameters $(s, t_x^c, t_y^c)$ with a neural network, where the scale parameter $s$ is converted to $t_z^c$ \cite{kanazawa_hmr,kanazawa_temporal_hmr,kocabas2019vibe,SPIN:ICCV:2019}. See \supmat~for details of this conversion.
The underlying assumption here is weak perspective projection. 
It is assumed that the camera is placed very far from the person, such that depth variations in the $z$ coordinate of the person are negligible compared to the distance from the camera. %as illustrated in Fig.~\ref{fig:naive_cam}(b).
This is violated regularly in natural images where it is common that the distance from the camera to the body is no more than the height of the body itself.

For intrinsic parameters, \cite{bogo_smplify,kocabas2019vibe,SPIN:ICCV:2019,SMPL-X:2019} set the focal length as a large constant $\focalx = \focaly = f \equiv 5000$ to meet the weak-perspective assumption and set the principal point $(\centerx, \centery)$ at the center of the resized cropped image around the person, while \cite{kanazawa_hmr,CenterHMR,zanfir2020weakly} directly apply weak perspective projection $x = sX + \camtransl$. %either a  $(\centerx, \centery)=(\cropwidth/2, \cropheight/2)$ .
Despite the differences in modeling projection and translation, one common feature of these simplified cameras is that they only have three unknowns: $(s, t_x^c, t_y^c)$ or equivalently $(t_x^c, t_y^c, t_z^c)$. 
We refer to them collectively as \emph{\iwcam} in this paper, standing for Identity rotation and Weak Perspective.

Note that both camera variables $(\camrot, \camtransl)$ and body variables $(\bodyori, \bodytransl)$ are expressed in world coordinates, not in camera space.   
Given just one single-view image, the network/optimizer can change both $(\camrot, \camtransl)$ and $(\bodyori, \bodytransl)$ to explain the image observations. 
\iwcam addresses this by assuming $\camrot = \text{I}$ and $\bodytransl=0$ to solve only for camera translation $\camtransl$ and body orientation $\bodyori$, or more precisely, the body orientation in the camera space: $\bodyori_c = \camrot \bodyori$.
See Fig.~\ref{fig:naive_cam}(a).
This approach is a key reason why most methods evaluate accuracy after Procrustes alignment of the estimated body to the ground truth.

\iwcam works well when the weak-perspective assumption holds. %as shown in  Fig.~\ref{fig:naive_cam}(c,d). 
However, images captured by cameras with significant pitch and smaller focal lengths, such as those in Fig.~\ref{fig:teaser} and Fig.~\ref{fig:qualitative}, 
have %strong 
foreshortening distortion that breaks this assumption because changes in the $z$ coordinate of the 3D body are no longer negligible compared to the distance from the camera.  
\iwcam expects HPS methods to absorb this camera pitch $\pitch$ into the relative body orientation $\bodyori_c$, but in practice, due to the mismatch in focal length, the optimization often unnecessarily changes body pose; e.g.~the wrong arm and leg poses in Fig.~\ref{fig:teaser}(b). %as shown in the ``wrong leg poses'' in Fig.~\ref{fig:teaser}(b).

\begin{SCfigure}
    \centering
    \includegraphics[width=0.25\textwidth]{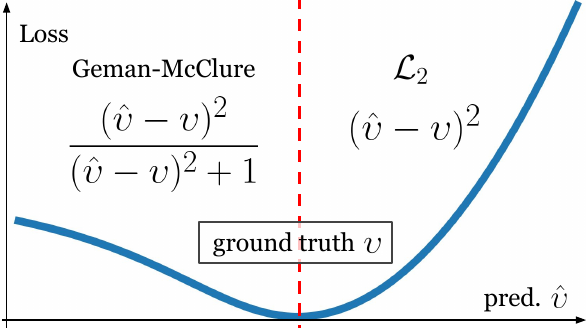}
    \caption{\softbiasedltwo penalizes underestimates of vfov less than over estimates. }%\textcolor{red}{See text.}}
    \label{fig:biased-L2}
    \vspace{-2ex}
\end{SCfigure}

\begin{figure*}[t]
    \centering
    \includegraphics[width=0.86\textwidth]{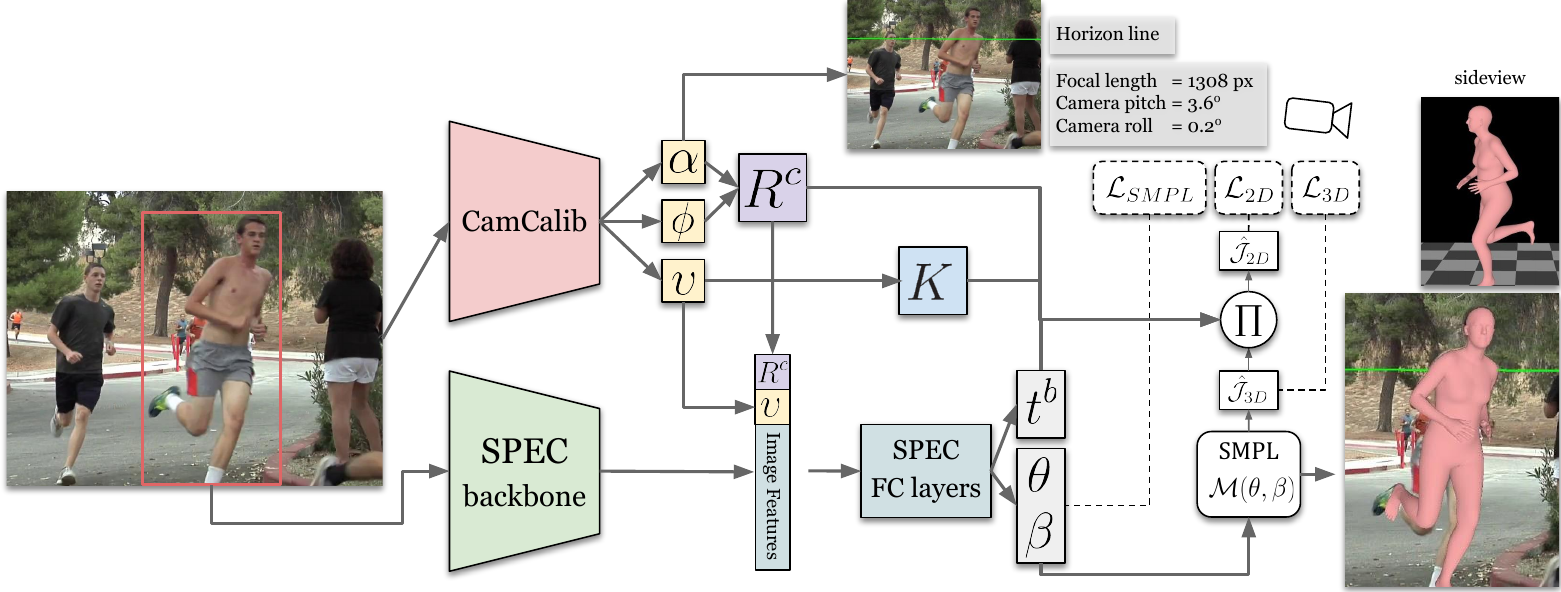}
    \caption{\textbf{\methodname overview.} \camcalib takes the whole input image as input and predicts camera pitch $\pitch$, roll $\roll$, and vertical field of view $\vfov$. These parameters are then used to construct camera rotation $\camrot$ and intrinsics $K$. Horizon line (green) is rendered following \cite{zhu2020single} to indicate camera rotations. \methodname takes a cropped bounding box as input and extracts image features using a CNN backbone. Predicted camera parameters from \camcalib are concatenated with image features to estimate SMPL body parameters $\theta,\beta$ along with the body translation $\bodytransl$. Camera parameters are also taken into account when computing a loss between the projected 3D joints $\predjoints{2}$ and ground truth.} 
    %\lea{While it is defined in the text, it would be nice to know from the Figure caption what $\pitch$, $\roll$,$\upsilon$ are.}}
    \label{fig:model_spec}
    \vspace{-2ex}
\end{figure*}{}
\subsection{Camera Calibration from a Single Image}
\label{cam_calib}
% \oh{This seems to be where the discription of our camera starts, we should dedicate at least a sub-heading to this.}

Inspired by single-view camera calibration and metrology \cite{Hold-Geoffroy_2018_CVPR,Kendall_2015_ICCV,workman2016horizon,zhu2020single}, we estimate camera rotation $\camrot$ and focal length $f$ from a single RGB image.
By lifting the zero camera rotation constraint, \ie~$\camrot \neq \text{I}$, and directly estimating $\camrot$, we bypass the camera-relative body orientation $\bodyori_c$, and thus disentangle camera rotations from body orientations.
Doing so allows us to handle perspective/foreshortening distortion 
while keeping a consistent $xz$-plane aligned ground plane located at $[0, y, 0]$.
Utilizing better focal length also improves the pose estimation quality by leveraging accurate perspective projection. 

Specifically, camera rotation is parameterized by three angles: pitch $\pitch$, roll $\roll$ and yaw. 
Since focal length in pixels has an unbounded range and it changes whenever one resizes images, we estimate vertical field of view (vfov) $\vfov$ in radians and convert it to focal length $f_y$ via:
\begin{equation}
\label{eq-vfov2fly}
    f_y = \frac{\frac{1}{2}h}{\tan (\frac{1}{2} \upsilon )},
\end{equation}
where $h$ is the image height in pixels.
% Without loss of generality, we assume zero yaw and $\focalx = \focaly = f $. 
We follow \cite{zhu2020single} to assume zero camera yaw and $\focalx = \focaly = f $.
% \textcolor{red}{We follow \cite{zhu2020single} to assume zero camera yaw and $\focalx = \focaly = f $.}
Our camera thus has three more parameters -- pitch, $\pitch$, roll, $\roll$, and vertical field of view, $\vfov$, in addition to the original $(s, t_x^c, t_y^c)$.
Since the three new parameters are all in radians, we choose to learn them with one camera calibration model termed \emph{CamCalib}. 
%leaving $(s, t_x^c, t_y^c)$ to the downstream body estimation tasks. 
We place the camera at the origin so $\camtransl = [0,0,0]$, as depicted in Fig.~\ref{fig:naive_cam}(b), and leave the estimation of body translation $\bodytransl$ to the downstream body estimator.

Many human body reconstruction networks take only a cropped image patch around the person as input.
In contrast, \camcalib takes the uncropped full-frame image to predict pitch $\pitch$, roll $\roll$, and vfov $\vfov$, which are the same for all subjects in the image.
We argue that this is beneficial because the full image contains rich cues that facilitate camera calibration. 
For example, there are abundant geometric cues, \eg~vanishing points and lines, available to help determine camera rotations and field of view in the original image. 
Following \cite{Hold-Geoffroy_2018_CVPR,Kendall_2015_ICCV,zhu2020single}, we use a CNN as the backbone for \camcalib and discretize the spaces of pitch $\pitch$, roll $\roll$, and vfov $\vfov$ into $B$ bins, converting the regression problem into a $B$-way classification problem.
However, instead of cross-entropy~\cite{workman2016horizon} or KL-divergence~\cite{Hold-Geoffroy_2018_CVPR,zhu2020single} losses,
we aggregate the predicted probability mass using a softargmax operation, \ie~computing the expectation value of the prediction, and measure its difference to the ground truth with an \ltwo loss which we term \softltwo. 
Thus, we avoid the difficulty of regressing in a continuous target space while retaining the notion of distance in the loss.
The detailed formulation of \softltwo is provided in the \supmat 

Furthermore, as pointed out by \cite{kissosECCVW2020,yu2020pcls} and shown in the \supmat, predicting larger focal lengths than the ground truth (or equivalently smaller fov) is less harmful to the reconstructed 3D poses than predicting smaller focal lengths (larger fov).
We therefore apply an asymmetric loss for vfov $\vfov$.
As shown in Fig.~\ref{fig:biased-L2}, 
predictions $\hat{\vfov}$ larger than the ground truth $\vfov$ yield higher penalty through a standard \ltwo loss, while the penalty for smaller predictions $\hat{\vfov}$ saturates via a Geman-McClure function \cite{geman-mcclure}. 
We verify the benefits of all these design choices in Sec.~\ref{experiments}.

\subsection{Optimization approach: \smplify}
\label{optimization_method}
Next, we showcase how using the estimated camera parameters help human body estimation in an optimization-based approach. 
To this end, we modify the SMPLify-X method \cite{SMPL-X:2019}. 
Given an image, \camcalib predicts camera pitch $\pitch$, roll $\roll$, and vfov $\vfov$.
We convert them into camera rotation $\camrot = R(\pitch)R(\roll)$ and intrinsics, $K$, storing $f=f_x=f_y$ and the principal point $(\centerx, \centery)=(w/2, h/2)$,
where $f_y$ is computed from Eq.~\ref{eq-vfov2fly} and $w,h$ are the image width and height in pixels. Then, we estimate the 2D keypoints $\gtjoints{2}$ using an off-the-shelf 2D keypoint detector~\cite{mmpose2020} and define the % \oh{we already refer to the intrinsic matrix in preliminaries. so either we assume that it is known to the reader and remove it here, or we move the definition to that section.}
the \smplify energy function as:
\begin{equation}
    E(\beta, \theta, \camrot, K, \bodytransl) = E_{J} + E_{\theta} + E_{\beta},
\end{equation}
where $\beta, \theta$ are SMPL shape and pose parameters, $\bodytransl$ is SMPL body translation, $E_{\theta}$ and $E_\beta$ are pose and shape prior terms, and $E_{J}$ is the data term. %$\bodytransl$, $\mathcal{M}$, $\Pi$ are the translation of SMPL body, the SMPL function, and the perspective projection, respectively. 
We modify the original SMPLify-X method to take perspective camera parameters into account in the data-term $E_{J}$. 
We obtain SMPL 3D joint locations using a pretrained joint regressor $W$ by $\predjoints{3} = W \mathcal{M}(\theta, \beta)$. 
$E_{J}$ measures the difference between the detected $\gtjoints{2}$ and the estimated $\predjoints{3}$, projected on the image by the estimated camera parameters:
\begin{equation}
\label{eq:persp-proj}
    E_{J} = \Big|\Big| \Pi \predjoints{3} - \gtjoints{2} \Big|\Big|^{2}_{2}\text{, where } \Pi = K[\camrot | -\bodytransl].
\end{equation}
\subsection{Learning-based approach: \methodname}
\label{regression_method}
To evaluate the effect of estimated camera parameters on a regression-based method, we take a simple and widely-used method, HMR \cite{kanazawa_hmr}, as a backbone, which employs a 2D reprojection loss during training that uses the estimated \iwcam. 
We incorporate our estimated camera parameters in two ways:
(1) by using them as $\camrot$ and $K$ during the projection of 3D joints like in Eq.~\ref{eq:persp-proj} and 
(2) by conditioning the fully connected layers, which estimate SMPL parameters, with $\camrot$ and $\vfov$.
Figure~\ref{fig:model_spec} gives an overview of \methodname.

Given an image, we first estimate the camera pitch $\pitch$, roll $\roll$, and vfov $\vfov$ using \camcalib and then convert them into $\camrot$ and $K$ as explained in Sec.~\ref{optimization_method}. %\smplify. %using the Eq.~\ref{eq:cam_int} and~\ref{eq:cam_rot}. 
For human body regression, we take a cropped bounding-box image as input and extract image features using a backbone CNN. 
These image features are concatenated with $\camrot$ and $\vfov$ and fed to an iterative regressor \cite{kanazawa_hmr} 
%\oh{we call this $W$ in the next paragraph -- might as well do that here} 
to regress SMPL pose $\theta$ and shape $\beta$ along with body translation $\bodytransl$. 
By doing so, \methodname learns to disentangle the SMPL body's global orientation $\bodyori$ from the camera rotation $\camrot$. See \supmat~for the details of $\bodytransl$.
Then, we obtain SMPL 3D joint locations $\predjoints{3}= W \mathcal{M}(\theta, \beta)$ and 2D projection $\predjoints{2}=\Pi \predjoints{3}$ as in Eq.~\ref{eq:persp-proj}.
Overall, our total loss for each training sample is:
$\mathcal{L} = \lambda_{3D}\mathcal{L}_{3D} + \lambda_{2D}\mathcal{L}_{2D} + \lambda_{SMPL} \mathcal{L}_{SMPL}$ where 
\begin{eqnarray}
\mathcal{L}_{\mathit{3D}} &= &  \| \predjoints{3} \; - \; \gtjoints{3} \|_F^2 \text{,  } \\
\mathcal{L}_{\mathit{2D}} &= & \| \predjoints{2} \; - \; \gtjoints{2} \|_F^2, \\
\mathcal{L}_{\mathit{SMPL}} &= & \| \hat{\theta} \; - \; \theta \|_2^2 + \| \hat{\beta} \; - \; \beta \|_2^2 \text{,}
\end{eqnarray}
where $\hat{\cdot}$ represents the prediction for the corresponding variable. %\oh{normally $\hat{\cdot}$ denotes the prediction. use $H$ as mnemonic aid - its often called the ``hat matrix'' that relates prediction to GT: $e=y-\hat{y}=y-Hy$. Or if you like puns: Why hat? Because, it is predicted to rain.} 
%\lea{$\hat{x}$ is used earlier already and I think it's clear without explaning it.}.
$\lambda$'s are scalar coefficients to balance the loss terms. %\oh{also these losses are computed over the entire training data so shouldn't we sum over (and normalize by) n}

Note that we define both 3D and 2D losses on the body joints.
This is because the 3D ground truth we use for training is not always reliable and thus the 2D joints provide important additional image cues, 
which can be exploited particularly well when using the correct camera geometry.

%% file: sections/4_experiments.tex
\section{Experiments}
\label{experiments}
We focus the evaluation on \camcalib and \methodname; for evaluation of \smplify, see \supmat

\subsection{Datasets}

\textbf{Pano360 dataset.} Previous work~\cite{Hold-Geoffroy_2018_CVPR, zhu2020single}, uses the SUN360~\cite{xiao2012recognizing} dataset to train camera calibration networks, which is unfortunately no longer available due to licensing issues. Therefore, we have curated a new dataset of equirectangular panorama images called Pano360. The Pano360 dataset consists of 35K panoramic images of which 34K are from Flickr and 1K rendered from photorealistic 3D scenes. Following previous work~\cite{Hold-Geoffroy_2018_CVPR, zhu2020single}, we randomly sample the camera pitch, roll, yaw, and vertical field of view to generate 400K training and 15K validation images. We use these to train our \camcalib model. %More details on the dataset  are provided in the \supmat

\textbf{\agoracam.} Existing datasets, used to train HPS regressors, contain limited camera variation. Hence, they are not ideal for training and evaluating the effects of camera estimation on HPS. 
Therefore, we created a photorealistic synthetic dataset, inspired by AGORA~\cite{patel2021agora}, to train and evaluate our model. 
It has high-quality textured human 3D scans and provides reference SMPL(-X) parameters for them. 
We place these scans in 5 different large high-quality photorealistic 3D scenes, enabling the generation of many unique views. 
We randomly sample camera viewpoints, $\pitch \sim \mathcal{U}(-30^{\circ}, 15^{\circ})$ and $\roll \sim \mathcal{N}(0^{\circ}, 2.8^{\circ})$, and focal lengths, $\vfov \sim \mathcal{U}(70^{\circ}, 130^{\circ})$, to add diversity.
In total, we generate 22191 images with 71982 ground truth bodies for training, and 3783 images with 12071 bodies for testing. 

\textbf{\mtpcam.} 
To evaluate calibrated HPS (CHPS) methods on real data, we collect a new dataset with high-quality pseudo ground truth using Amazon Mechanical Turk (AMT).
%\mtpcam is a new test set for human pose, shape, and camera estimation. 
Following the idea of the MTP dataset~\cite{Mueller:CVPR:21}, we ask AMT workers to mimic 10 poses for which we have 3D ground truth. 
While the person maintains a pose, a second person records a video from different viewpoints.
%: chest height, overhead, and close to the ground.
In addition, the worker calibrates the camera
%, following a detailed protocol 
and provides their height and weight.
We extend SMPLify-XC~\cite{Mueller:CVPR:21} and use the calibrated camera to fit the SMPL-X model to multiple video frames. See \supmat~for details.
In total, we collect 64 videos of 7 subjects (4 male, 3 female) and extract 3284 
%\lea{replace with actual size of test set} 
images at a frame rate of 1 fps. 

\textbf{Other datasets.} 
To train 3D CHPS estimation, we use the 3DPW~\cite{vonMarcard2018_3dpw}, COCO~\cite{coco}, MPI-INF-3DHP~\cite{mpiiinf3dhp_mono-2017}, and Human3.6M~\cite{ionescu_h36m} datasets. 
We evaluate \methodname using separate test data: 3DPW-test, \agoracam, and \mtpcam. 
Since there are no ground truth 3D body and camera annotations for COCO, we use \camcalib to estimate camera parameters and SMPLify-X-cam to obtain pseudo 3D body ground truth, using EFT~\cite{joo2020eft} annotations as the initialization.

\subsection{Evaluation metrics} 
\label{sec:metrics}
The mean per joint position error (\mpjpe), Procrustes-aligned mean per joint position error (\pampjpe), and per vertex error (\pve) are the most commonly-used evaluation metrics in the literature. 
PA-MPJPE exists as a metric specifically because current HPS methods that use \iwcam reconstruct bodies in camera coordinates $\predjoints{3}^{cam}$. Procrustes alignment ``hides many sins" in that it removes the rotation of the body caused by an unknown camera pose. 
See \supmat~for details of how PA-MPJPE and MPJPE are computed.

Instead, we propose variants of \mpjpe and \pve that compute the error in world coordinates without the need of camera information and dub them \wmpjpe and \wpve. 
Since \methodname disentangles camera and body rotations, the predictions reside in world coordinates $\predjoints{3}^{world}$ 
% thanks to the use of estimated cameras from \camcalib, 
and \wmpjpe is computed as $\lVert \predjoints{3}^{world} - \gtjoints{3} \rVert$. 
For existing SOTA methods, we report two versions of \wmpjpe: 
(1) $\lVert \predjoints{3}^{cam} - \gtjoints{3} \rVert$
and,
(2) $\lVert R^{c^{-1}} \predjoints{3}^{cam} - \gtjoints{3} \rVert$, where $\camrot$ is the estimated camera rotation by \camcalib.
By reporting (2), we do not assume known camera rotations for any method and compare them all in world coordinates. 
This also illustrates the effect of using \camcalib with prior work. 
% These evaluation metrics are discussed in greater detail in \supmat \textcolor{red}{We also report and discuss MPJPE \& PVE in the \supmat}
We discuss these metrics in greater detail and report MPJPE \& PVE in the \supmat

\begin{table}[]
    \centering
    \resizebox{0.48\textwidth}{!}{
        \npdecimalsign{.}
        \nprounddigits{2}
        \begin{tabular}{l|r|r|r}
            \toprule
            \textbf{Methods} & \textbf{vfov $\vfov^{\circ}$} $\downarrow$ & \textbf{pitch $\pitch^{\circ}$} $\downarrow$ & \textbf{roll $\roll^{\circ}$} $\downarrow$ \\ \midrule
            ScaleNet~\cite{zhu2020single} & 5.68 & 2.61 & 1.41 \\ \midrule
            \camcalib (KL Loss) & 3.53 & 2.32 & 1.15 \\ %\midrule
            \camcalib (\softltwo) & 3.34 & 2.06 & 1.11 \\ %\midrule
            \camcalib (\softbiasedltwo) & \bf 3.24 & \bf 1.94 & \bf 1.02 \\
            \bottomrule
        \end{tabular} 
    }
    \vspace{-0.05in}
    \caption{{\bf Regressing camera parameters.} \camcalib methods are trained and tested on the Pano360 dataset. ScaleNet~\cite{zhu2020single} results use the authors' implementation.
    }
    \vspace{-2ex}
    \label{tab:camcalib}
\end{table}

\subsection{Implementation details}
\label{sec:implementation}

\textbf{\camcalib.} We follow the implementation of \cite{Hold-Geoffroy_2018_CVPR,zhu2020single} but use ResNet-50 as the backbone and predict pitch $\pitch$, roll $\roll$, and vfov $\vfov$ with separate fully-connected layers. 
Each parameter has $B=256$ bins and we apply \softbiasedltwo for $\vfov$, \softltwo for $\pitch$ and $\roll$.
The model is trained with images of varied resolutions for 30 epochs. The Pano360 dataset is used for training and evaluation.

\textbf{\methodname.} 
Similar to the original HMR~\cite{kanazawa_hmr}, we use a ResNet-50 backbone, followed by fully-connected layers that iteratively regress SMPL parameters. 
We do not apply HMR's adversarial discriminator since we use psuedo-ground-truth 3D training data. 
The Adam optimizer with a learning rate of $5e^{-5}$ is used. 
For the first 150 training epochs we use the COCO and \agoracam datasets, and then incorporate MPI-INF-3DHP and Human3.6M. 
Total training takes around 175 epochs, $\sim$4-5 days.

Note that \camcalib and \methodname are trained separately. 
Training them jointly is not possible because we lack an \emph{in-the-wild} dataset that has both ground-truth bodies and diverse camera focal lengths and views.
\mtpcam meets these requirements but is small so we us it for evaluation. During inference, CamCalib and SPEC run jointly.

\subsection{Single-image camera calibration results}
Table~\ref{tab:camcalib} reports the mean angular error in camera pitch $\pitch$, roll $\roll$, and vfov $\vfov$ for different camera calibration methods on the Pano360 test set.
For reference, we inlcude the open-source implementation of ScaleNet \cite{zhu2020single}, which uses a different backbone than \camcalib.
We also train ScaleNet with our backbone on Pano360 (\camcalib (KL loss) in Table~\ref{tab:camcalib}).
%The architecture of \camcalib and ScaleNet is the same but ScaleNet uses a KL loss for training. 
We evaluate the effect of different loss functions, \ie~KL-divergence, \softltwo, and \softbiasedltwo, and define the final \camcalib network to be the best performing version (\softbiasedltwo).
% In Figure~\ref{fig:cam_calib_results}, we show the results of a CNN model trained to estimate perspective camera parameters.

% \subsection{CamCalib results}

%%%%%%%%%%%%%%%%%%%%%%%%%%%%%%%%%%%%%%%%%%%%%%%%%%%%%%%%%%%%%%

\subsection{\methodname evaluation}
Tables~\ref{tab:sota_mtp}, \ref{tab:sota_agora}, and \ref{tab:sota_3dpw} show the results of recent SOTA methods on the \mtpcam, \agoracam and 3DPW datasets. 

{\bf The correct metric.} 
We believe W-MPJPE is the metric that best reflects performance in real-world applications, so we report \wmpjpe, \pampjpe, and \wpve. %\eaksan{why not MPJPE (in the camera space)?}
\mpjpe is also reported and discussed in \supmat
For \wmpjpe and \wpve, we report both definitions from Sec.~\ref{sec:metrics}, i.e.~(1)/(2) in the tables.
Note that \methodname has the same error under both metrics.
\pampjpe has only one entry because Procrustes alignment removes the effect of camera rotation (and more); this effectively hides the fact that SOTA methods do not estimate global pose well.

{\bf Comparison to the state-of-the-art.} 
To compute the performance of SOTA methods, we use their open source implementations. 
We use HMR$^{*}$ as our \iwcam baseline, 
which is HMR~\cite{kanazawa_hmr} trained with the same datasets as \methodname, \ie~COCO, \agoracam, MPI-INF-3DHP, and Human3.6M. 
Again, we do not use HMR's discriminator since we train with ground-truth or pseudo ground-truth 3D labels. %Therefore we denote  
For I2L-MeshNet~\cite{Moon_2020_ECCV_I2L-MeshNet}, we use the SMPL output of this method instead of non-parametric mesh to be able to report \wpve and denote this with $^{\dagger}$.

\begin{figure*}[h]
    \centering
    \includegraphics[width=0.87\textwidth]{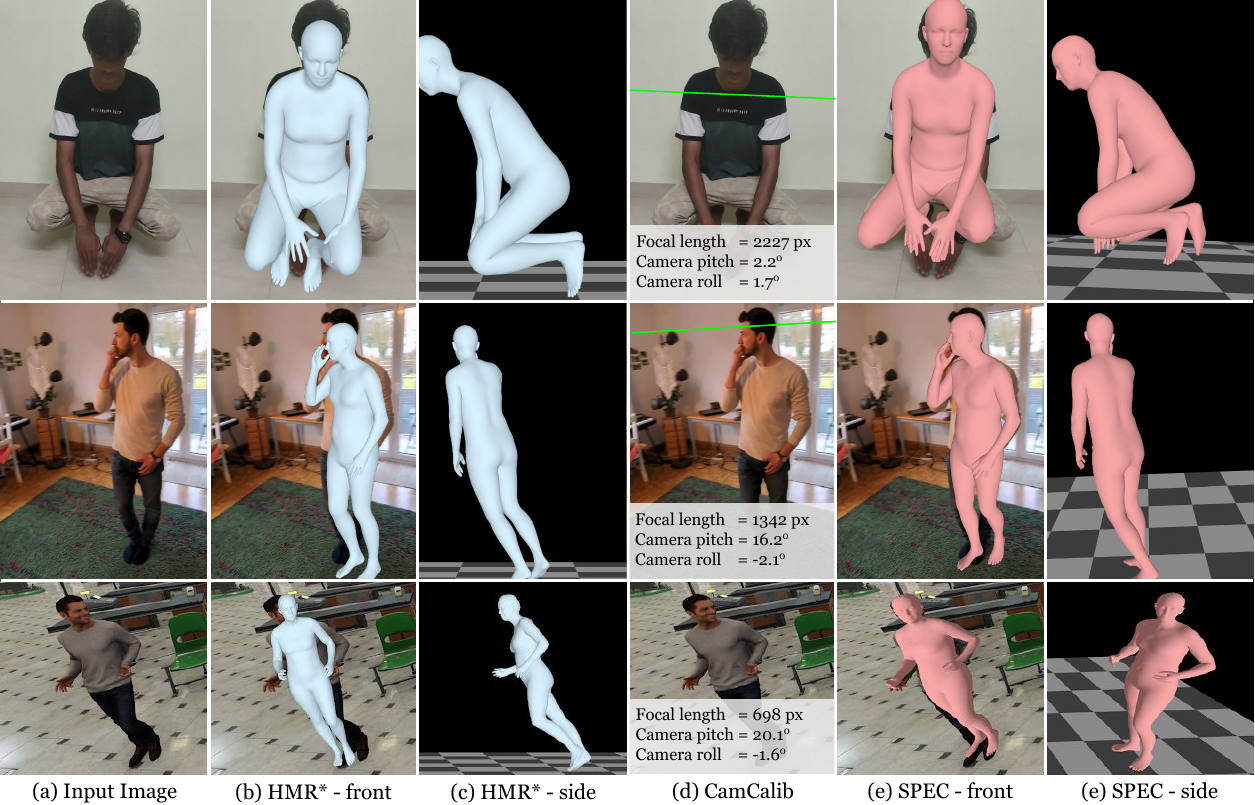}
    \vspace{-0.1in}
    \caption{Qualitative results. Top \& middle: \mtpcam; bottom: \agoracam. We also provide failure cases in \supmat 
    % \mkocabas{I will clean the green line and top burn-in for the input image column. Add camcalib predictions as a small textbox for each image for the 4th column.} 
    }
    \label{fig:qualitative}
    \vspace{-2ex}
%    \vspace{-2ex}
\end{figure*}

Since \wmpjpe and \wpve measure the error w.r.t.~the body in world coordinates, 
these measures reveal the performance improvement of \methodname over the SOTA when the camera deviates from the \iwcam assumption. 
Compared a dataset like 3DPW, \mtpcam and \agoracam have significantly more variety in focal lengths and viewpoints as shown in Fig.~\ref{fig:qualitative}.
As a result, \methodname yields a larger improvement in \wmpjpe and \wpve  over the SOTA for these datasets.
Using explicit camera information is a key driver of this improvement. 
The improvement in \pampjpe is less significant, suggesting that the largest improvements come from estimating the body in world coordinates rather than better articulated pose.
This is valuable in many applications, \eg~human-scene interaction, where bodies and objects are often reconstructed from distinct methods but should reside in a common space.

\begin{figure}[h]
    \centering
    \begin{minipage}{.25\textwidth}
        \centering
        \includegraphics[width=0.95\textwidth]{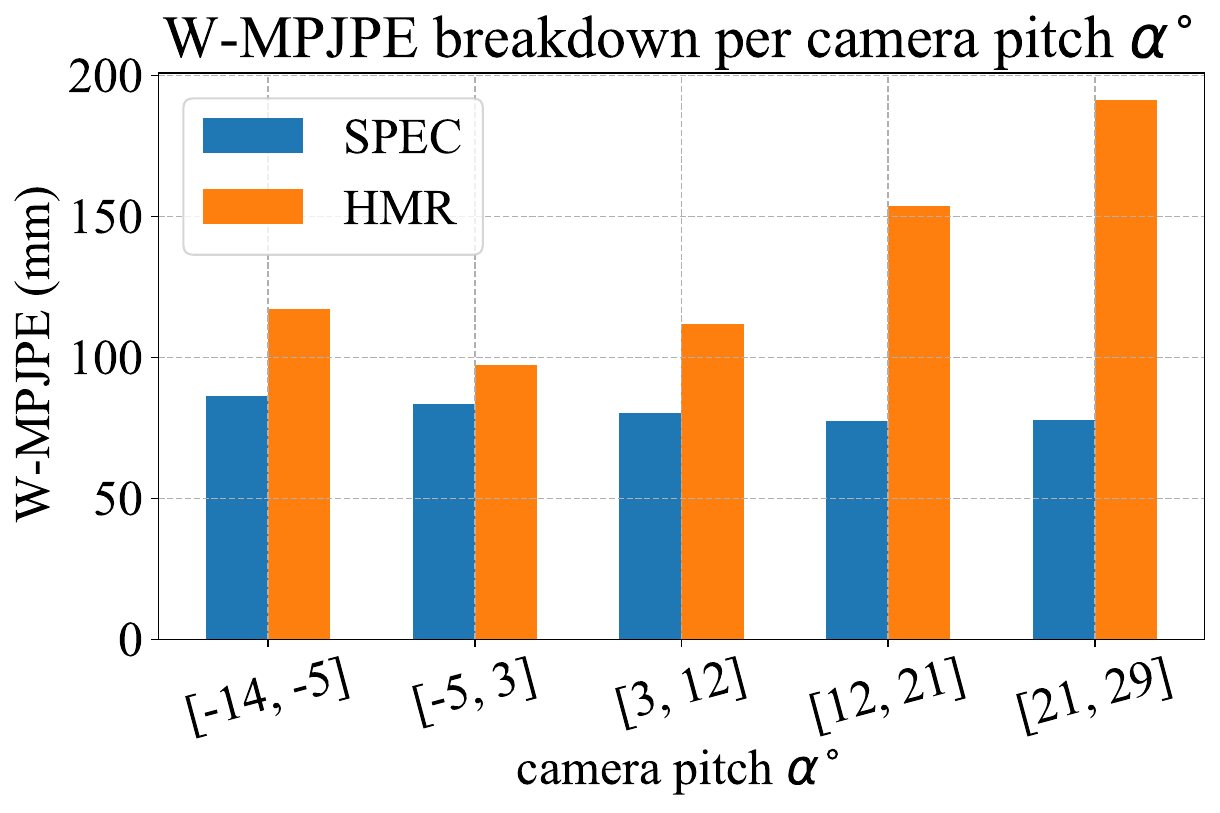}
        % \captionof{figure}{A figure}
        % \label{fig:test1}
    \end{minipage}%
    \begin{minipage}{.25\textwidth}
        \centering
        \includegraphics[width=0.95\textwidth]{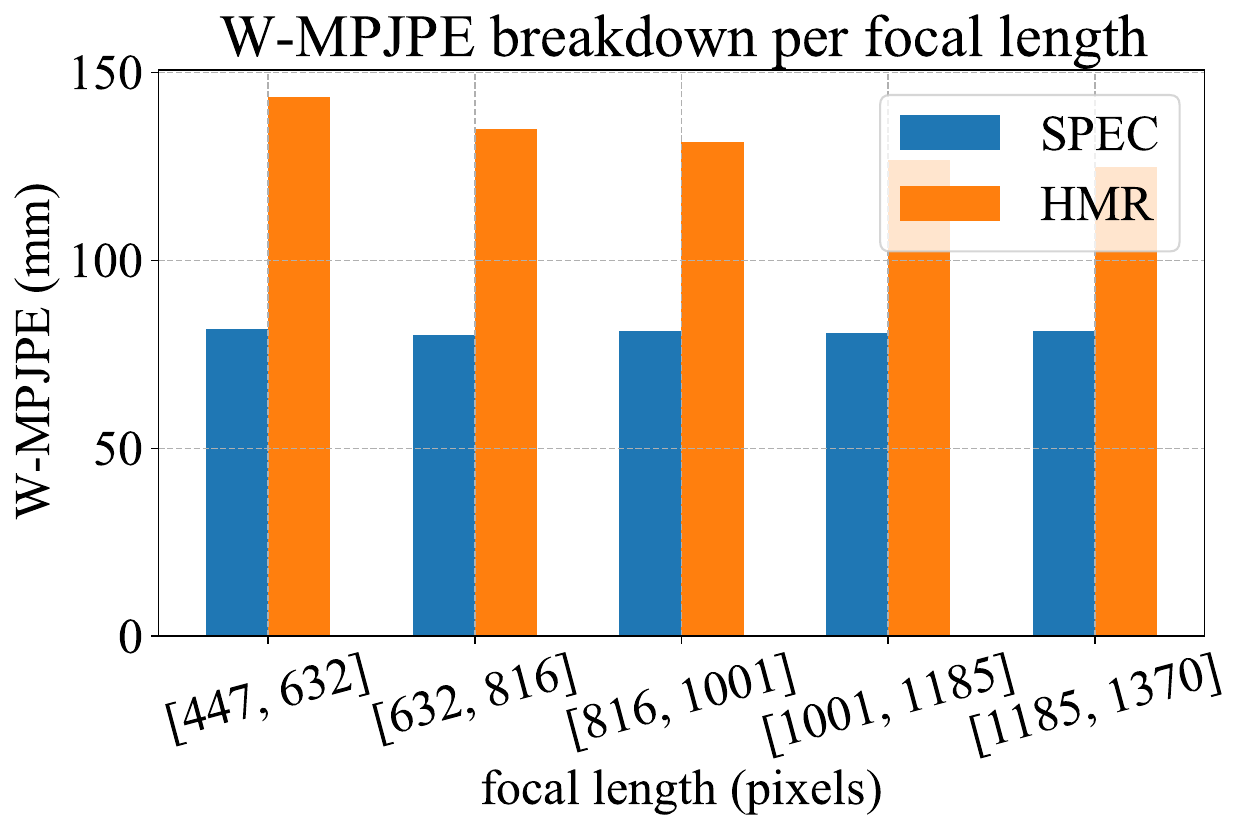}
        % \captionof{figure}{A figure}
        % \label{fig:test1}
    \end{minipage}%
    \vspace{-0.1in}
    \caption{Breakdown of \wmpjpe per camera focal length and pitch range.}
    \label{fig:error_breakdown}
    \vspace{-3ex}
\end{figure}

Figure \ref{fig:error_breakdown} analyzes \wmpjpe on \agoracam for different camera viewpoints and focal lengths. \methodname results are similar across different camera settings, while the HMR$^{*}$ is less robust to values that deviate from its assumptions.

\input{tables/results_mtp_j24}

\input{tables/results_agora_j24}

\input{tables/results_3dpw}

\input{tables/results_agora_ablation_j24}

\textbf{Ablation experiments.} We ablate different camera parameters and components of \methodname to investigate their effect on performance. 
We report results on the \agoracam test data (Table \ref{tab:hmr_agora}) because it has challenging cameras; for ablation results on 3DPW please see \supmat %\oh{weird double punctuation; must be the macro}

We use HMR$^{*}$ as our baseline and use the same datasets and training configurations for all methods. 
To obtain the predicted 2D joints, $\predjoints{2}$, HMR$^{*}$ uses the bounding-box center as the principal point. 
We start by changing it to the image center, \ie~$o_x=w/2, o_y=h/2$ denoted as ``HMR$^{*}$ + $c$". 
This ensures a better projective geometry than using the bounding-box center as the image center and already improves results. 
Next, we replace the fixed focal length of $5000$ with the values estimated by \camcalib, denoted as ``HMR$^{*}$ + $c$ + $f$". 
``HMR$^{*}$ + $c$ + $f$ + $\camrot$" uses $\camrot$ instead of the identity matrix as the camera rotation during projection. 
Finally, \methodname uses $c$, $f$, and $\camrot$ both during projection and as a conditioning input to the final stage of HMR$^{*}$ predictions. 
Overall, improving the camera model improves  \wmpjpe and \wpve. 
Conditioning the network on the camera parameters helps, but we suspect that better camera conditioning schemes can be employed to make the network more aware of the camera geometry. 

\textbf{Qualitative results.}
Figure \ref{fig:qualitative} shows representative results.
HMR$^{*}$ assumes \iwcam and yields incorrect body poses (legs in top row) and incoherent body orientations (middle \& bottom);
\methodname predicts overall more globally coherent bodies as can be seen in side-view images. 

%% file: tables/results_mtp_j24.tex
\begin{table}
    \centering
    \resizebox{0.47\textwidth}{!}{
        \begin{tabular}{l|r|r|r}
        \toprule
        \textbf{Methods} & \textbf{\wmpjpe} & \textbf{\pampjpe} & \textbf{\wpve} \\ 
        \midrule 
        GraphCMR~\cite{kolotouros2019cmr} & 175.1 / 166.1 & 94.3 & 205.5 / 197.3 \\
        SPIN~\cite{SPIN:ICCV:2019} & 143.8 / 143.6 & 79.1 & 165.2 / 165.3 \\
        % Pose2Mesh~\cite{Choi_2020_ECCV_Pose2Mesh} & & & & & \\
        PartialHumans~\cite{Rockwell2020} & 158.9 / 157.6 & 98.7 & 190.1 / 188.9 \\
        I2L-MeshNet$^{\dagger}$~\cite{Moon_2020_ECCV_I2L-MeshNet} & 167.2 / 167.0 & 99.2 & 199.0 / 198.1 \\
        % \rowcolor{pink} PARE & 143.3 / 127.8 & 69.7 & 160.7 / 146.5 \\
        % CenterHMR~\cite{CenterHMR} & 185.5 & 70.3 & 197.4 \\
        % CRMH~\cite{jiang2020mpshape} & & & & & \\ 
        \midrule
        HMR$^{*}$~\cite{kanazawa_hmr} & 142.5 / 128.8 & \textbf{71.8 }& 164.6 / 150.7\\
        \methodname & \textbf{124.3} / \textbf{124.3} & \textbf{71.8} & \textbf{147.1} / \textbf{147.1}\\
        \bottomrule
        \end{tabular} 
    }
    \vspace{-0.05in}
    \caption{Results of SOTA methods on \mtpcam dataset. 
    We use the implementations provided by the authors to obtain results. 
    HMR$^{*}$ means that we train HMR using the same data as \methodname for fair comparison. $^{\dagger}$means we use the SMPL output of this method instead of the non-parametric mesh to be able to report \wpve. All numbers are in \emph{mm}. % \mkocabas{Evaluation protocol is updated to use 24 joints.}
    %We use the implementations provided by authors to obtain results. HMR$^{*}$ denotes that we train HMR using the same data as \methodname to make a fair comparsion. $^{\dagger}$ we use the SMPL output of this method instead of non-parametric mesh to be able to report \wpve.
    }
    \vspace{-2ex}
 %   \vspace{-1ex}
    \label{tab:sota_mtp}
\end{table}

%% file: tables/results_agora_j24.tex
\begin{table}
    \centering
    \resizebox{0.47\textwidth}{!}{
        \begin{tabular}{l|r|r|r}
        \toprule
        \textbf{Methods} & \textbf{\wmpjpe} & \textbf{\pampjpe} & \textbf{\wpve} \\ 
        \midrule 
        GraphCMR~\cite{kolotouros2019cmr} & 181.7 / 181.5 & 86.6 & 219.8 / 218.3 \\
        SPIN~\cite{SPIN:ICCV:2019} & 165.8 / 161.4 & 79.5 & 194.1 / 188.0 \\
        % Pose2Mesh~\cite{Choi_2020_ECCV_Pose2Mesh} & & & & & \\
        PartialHumans~\cite{Rockwell2020} & 169.3 / 174.1 & 88.2 & 207.6 / 210.4 \\
        I2L-MeshNet$^{\dagger}$~\cite{Moon_2020_ECCV_I2L-MeshNet} & 169.8 / 163.3 & 82.0 & 203.2 / 195.9 \\
        % \rowcolor{pink} PARE & 164.2 / 148.1 & 69.4 & 189.2 / 172.1 \\
        % CenterHMR~\cite{CenterHMR} & 185.5 & 70.3 & 197.4 \\
        % CRMH~\cite{jiang2020mpshape} & & & & & \\ 
        \midrule
        HMR$^{*}$~\cite{kanazawa_hmr} & 128.7 / 96.4 & 55.9 & 144.2 / 111.8 \\
        \methodname & \textbf{74.9} / \textbf{74.9} & \textbf{54.5} & \textbf{90.5} / \textbf{90.5} \\
        \bottomrule
        \end{tabular} 
    }
    \vspace{-0.05in}
    \caption{Results of SOTA methods on \agoracam. See Table~\ref{tab:sota_mtp} caption. 
    % \mkocabas{Evaluation protocol is updated to use 24 joints.}
    }
    \vspace{-3ex}
%    \vspace{-1ex}
    \label{tab:sota_agora}
\end{table}

%% file: tables/results_3dpw.tex
\begin{table}
    \centering
    \resizebox{0.47\textwidth}{!}{
        \begin{tabular}{l|r|r|r}
        \toprule
        \textbf{Methods} & \textbf{\wmpjpe} & \textbf{\pampjpe} & \textbf{\wpve} \\ 
        \midrule 
        % GraphCMR~\cite{kolotouros2019cmr} & 198.324 & 197.182 & 79.779 & 215.141 & 227.888 \\
        GraphCMR~\cite{kolotouros2019cmr} & 137.8 / 129.4 & 69.1 & 158.4 / 152.1 \\
        SPIN~\cite{SPIN:ICCV:2019} & 122.2 / 116.6 & 59.0 & 140.9 / 135.8 \\
        Partial Humans~\cite{Rockwell2020}& 139.4 / 132.9 & 76.9 & 160.1 / 152.7 \\
        I2L-MeshNet$^{\dagger}$~\cite{Moon_2020_ECCV_I2L-MeshNet} & 133.3 / 119.6 & 60.0 & 154.5 / 141.2 \\
        % CenterHMR~\cite{CenterHMR} & 130.3 & 60.8 & 149.3 \\
        \midrule
        HMR$^{*}$~\cite{kanazawa_hmr}  & 119.2 / \textbf{104.0} & 53.7 & 136.2 / \textbf{120.6}\\
        \methodname & \textbf{106.4} / 106.4 & \textbf{53.2} & \textbf{127.4} / 127.4 \\ 
        \bottomrule
        \end{tabular} 
    }
     \vspace{-0.05in}
   \caption{Results of SOTA methods on 3DPW test set. See Table~\ref{tab:sota_mtp} caption.
    %We use the implementations provided by authors to obtain results. HMR$^{*}$ denotes that we train HMR using the same data as \methodname to make a fair comparsion. $^{\dagger}$ we use the SMPL output of this method instead of non-parametric mesh to be able to report \wpve.
    }
    \vspace{-2ex}
%    \vspace{-1ex}
    \label{tab:sota_3dpw}
\end{table}

%% file: tables/results_agora_ablation_j24.tex
\begin{table}
    \centering
    \resizebox{0.47\textwidth}{!}{
        \begin{tabular}{l|r|r|r}
        \toprule
        \textbf{Methods} & \textbf{\wmpjpe} & \textbf{\pampjpe} & \textbf{\wpve} \\ 
        \midrule
        
        HMR$^{*}$ & 128.7 / 96.4 & 55.9 & 144.2 / 111.8\\
        HMR$^{*}$ + $c$ & 120.4 / 84.2 & 54.0 & 135.3 / 98.8\\
        HMR$^{*}$ + $c$ + $f$ &  118.3 / 85.1 & 54.0 & 132.8 / 99.7 \\ 
        \midrule
        HMR$^{*}$ + $c$ + $f$ + $\camrot$ & 77.2 / 77.2 & 55.3 & 93.8 / 93.8 \\ 
        \methodname & \textbf{74.9} / \textbf{74.9} & \textbf{54.5} & \textbf{90.5} / \textbf{90.5} \\
        \bottomrule
        \end{tabular} 
    }
    \vspace{-0.05in}
    \caption{Ablation studies with \agoracam. $c$: using the image center as camera center. 
    % $f$: using \camcalib estimated focal length, $\camrot$: using \camcalib estimated camera rotation.
    $f$ and $\camrot$: using \camcalib estimated focal length and camera rotation, respectively.
    }
    \vspace{-2ex}
%    \vspace{-3ex}
    \label{tab:hmr_agora}
\end{table}

%% file: sections/5_conclusion.tex
\section{Conclusion}
\label{conclusion}
In this paper, we demonstrate that a) camera geometry can be estimated from images and b) can effectively be leveraged to improve 3D HPS accuracy. 
Existing methods make simplifying assumptions about the camera: weak-perspective projection, large constant focal length, and zero camera rotation. 
To go beyond these simple assumptions, we introduce \methodname, the first 3D HPS method that regresses a perspective camera from a single image and employs this to reconstruct 3D human bodies more accurately.
Using the estimated camera parameters improves both SOTA camera regression methods and HPS regression methods.
We introduce two new datasets, \ie~\mtpcam and \agoracam, with accurate camera and 3D body annotations to showcase the effect of \methodname through ablation studies and comparison with the SOTA and to foster future research in this area.

%% file: sections/8_acknowledgements.tex
\small{
\noindent
{\bf Acknowledgements:}
We thank Emre Aksan, Shashank Tripathi, Vassilis Choutas, Yao Feng, Priyanka Patel, Nikos Athanasiou, Yinghao Huang, Cornelia Kohler, Hongwei Yi, Dimitris Tzionas, Nitin Saini, and all Perceiving Systems department members for their feedback and the fruitful discussions. This research was partially supported by the Max Planck ETH Center for Learning Systems.

\noindent{\bf Disclosure:} \url{https://files.is.tue.mpg.de/black/CoI/ICCV2021.txt}
}

%% file: sections/9_supmat.tex
\section{Methods}
\subsection{Formulation of per-body translation $\bodytransl$}

For each body in the image, besides SMPL parameters $\theta,\beta$, \methodname also estimates camera parameters $(s, t_x, t_y)$, 
which is defined w.r.t.~the bounding box (bbox) of the subject.
Similar to \cite{jiang2020mpshape,kissosECCVW2020}, we perform a coordinate transformation 
to obtain the final $\bodytransl$ vector w.r.t.~the original full image following: 
\begin{equation}
\label{eq:body_transl}
\begin{split}
    \bodytransl_x & = t_x + \frac{2(\cropcenterx - \imgwidth/2)}{s \cdot \cropwidth} \text{,} \\
    \bodytransl_y & = t_y + \frac{2(\cropcentery - \imgheight/2)}{s \cdot \cropheight} \text{, and } \\
    \bodytransl_z & = \frac{2 \cdot f}{\cropheight \cdot s} \text{,}
\end{split}
\end{equation}
where $(\cropcenterx, \cropcentery)$ is the bbox center, $\imgwidth, \imgheight$ are originial image sizes, $\cropwidth, \cropheight$ are bbox sizes, and focal length $f$ is estimated by \camcalib.
As explained in the main text, Eq.~3, $\bodytransl$ is used during perspective projection $\Pi = K[\camrot|-\bodytransl]$.

\subsection{\softltwo and \softbiasedltwo}
As described in the main manuscript, we follow \cite{zhu2020single} to discretize the spaces of pitch $\pitch$, roll $\roll$ and vertical field of view (vfov) $\vfov$ into $B=256$ bins  but avoid casting it as a pure classification problem.
To this end, we propose \softltwo loss and the biased variant.
Let
$\mathbf{\boldsymbol{\pitch}}=\left [ \pitch_1, \dots \pitch_i, \dots \pitch_B \right]$,
$\mathbf{\boldsymbol{\roll}}=\left [ \roll_1, \dots \roll_i, \dots \roll_B \right]$, and 
$\mathbf{\boldsymbol{\vfov}}=\left [ \vfov_1, \dots \vfov_i, \dots \vfov_B \right]$ denote the center values of each of the bins, 
and let 
$\mathbf{p^{\pitch}}=\left [ p_1^{\pitch}, \dots p_i^{\pitch}, \dots p_B^{\pitch} \right]$,
$\mathbf{p^{\roll}}=\left [ p_1^{\roll}, \dots p_i^{\roll}, \dots p_B^{\roll} \right]$, and
$\mathbf{p^{\vfov}}=\left [ p_1^{\vfov}, \dots p_i^{\vfov}, \dots p_B^{\vfov} \right]$ 
denote the probability mass from the fully-connected layers of each head respectively.
We compute the expectation value of the probability mass as the prediction:
\begin{equation}
\begin{split}
    \hat{\pitch} &= \sum_i p_i^{\pitch} \pitch_i\text{,}  \\ 
    \hat{\roll} &= \sum_i p_i^{\roll} \roll_i \text{, and} \\ 
    \hat{\vfov} &= \sum_i p_i^{\vfov} \vfov_i. \\ 
\end{split}
\end{equation}
This differentiable operation has been commonly-used in human joint detection \cite{luvizon2019human,sun2018integral} 
to determine the peak location in a likelihood heat map, in contrast to the non-differentiable argmax operation.

For pitch $\pitch$ and roll $\roll$ angles, we apply the standard $\mathcal{L}_2$ loss between the prediction and the ground truth. 
To encourage underestimation of vfov $\vfov$ more than overestimation, 
we design an asymmetric loss as depicted in Fig.~3 in the main text; formally:
\begin{equation}
\label{eq-softbiasedltwo}
\mathcal{L}(\hat{\vfov})=
    \begin{cases}
       \frac{(\hat{\vfov} - \vfov)^2}{(\hat{\vfov} - \vfov)^2+ 1}, & \text{if}\ \hat{\vfov} <= \vfov \\
      (\hat{\vfov} - \vfov)^2, & \text{otherwise.}
    \end{cases}
\end{equation}
We verify the benefits of these design choices in Table 1 in the main text.

\subsection{Virtual ground plane}
In all qualitative results, we visualize a virtual ground plane with a checkerboard, 
which is parallel to the $xz$-plane and therefore parameterized as $[0,y,0]$.
We define $y=\min\left(\mathcal{M}(\theta, \beta)[:,2]\right)$, \ie~we place the ground plane just below the SMPL mesh. 

This simple parameterization is feasible because we disentangle the camera rotation $\camrot$ from the body orientation $\bodyori$.
For SOTA methods that apply the \iwcam model, the virtual ground planes are often tilted.
As a result, it requires further processing to estimate the up-vector, or conversely, the direction of gravity, 
making it non-trivial to integrate the reconstructed bodies for some downstream applications, \eg~scene understanding, character animation, physics simulation. 
\methodname, on the other hand, reconstructs bodies in the world coordinate frame with a consistent up vector $[0,1,0]$, 
which is more physically plausible when visualized together with the ground plane. See \supmat~video.

\subsection{SMPLify-X-Cam} 
To integrate the estimated camera parameters from \camcalib into an optimization-based method, 
we use the original implementation of SMPLify-X \cite{SMPL-X:2019} with slight modifications. 
We replace the \iwcam with the estimated $K$ and $\camrot$ as described in Sec.~3.3 of the main text.
Additionally, we initialize the optimization with the output of HMR-EFT \cite{joo2020eft}, instead of starting from a mean pose and a mean shape.
We use the Adam optimizer with a step size of $10^{-2}$ for 100 steps for both the first and second stage of optimization.
The results of \smplify are evaluated in Sec.~\ref{sec:smplifycam_evaluation}.

\section{Implementation Details}
\subsection{\mtpcam Dataset}
M\"uller \etal~\cite{Mueller:CVPR:21} propose a ``Mimic The Pose'' (MTP) task 
to collect datasets of natural images with high-quality pseudo ground truth body parameters from Amazon Mechanical Turk (AMT).
Each image shows a person mimicking a posed SMPL-X mesh presented to them on AMT. 
M\"uller \etal~devise a three-stage optimization routine, \smplifyxc, to fit the SMPL-X model to each image. 
It is based on the default SMPLify-X \cite{SMPL-X:2019} but considers additionally the pose $\tilde{\theta}$ and self-contact $\tilde{C}$ of the presented SMPL-X mesh to constrain the optimization. 
The body parts in self-contact are identified by finding vertices that are close in Euclidean and far in geodesic space. 
Please see \cite{Mueller:CVPR:21} for a detailed definition of self-contact.

We follow the MTP approach but add two distinctions in order to obtain ground truth camera parameters:
(1) Instead of getting a single picture for each pose, we ask AMT subjects to record a video showing the pose from multiple viewpoints, similar to Mannequin-Challenge style \cite{li2019learning}.
(2) We also ask them to print out a calibration pattern and record a video of the grid following a detailed protocol. In addition, we ask them to measure the size of the grid and take a picture of the grid and a ruler to verify the measured values.

To fit SMPL-X pose $\theta$ and shape $\beta$, as well as camera pitch $\pitch$, roll $\roll$, yaw $\yaw$, 
and camera translation $\camtransl$ to the collected MTP videos, we extend \smplifyxc and introduce SMPLify-XC-Cam. 
We follow the three-stage optimization routine.
In the first stage, body pose $\theta$ is initialized as the poses of the presented mesh, $\theta = \tilde{\theta}$, and stays fixed in this stage. The objective is:
\begin{equation}
E(\beta, \pitch, \roll, \yaw, \camtransl) = \lambda_M E_M + \sum_{i=1}^{F} E_{J_i} \text{.}
\end{equation}
$E_M$ denotes the SMPLify-XC shape loss, that takes the ground truth height and weight of a person into account. 
\begin{equation}
    E_{J_i} = \Big|\Big| \omega_i \gamma \big( \Pi \big(\mathcal{M}(\beta, \theta), \camrot_i, K, \camtransl_{i} \big) - \mathcal{J}_{\mathit{2D}_i} \big) \Big|\Big|^{2}_{2}
\end{equation}
is the 2D re-projection error of a single frame $i$ with detected 2D joints $\mathcal{J}_{\mathit{2D}_i}$, and camera rotation $\camrot_i$ and translation $\camtransl_i$. $\omega_i$ and$\gamma$ are per joint confidence and weight, respectively. $F$ is the number of frames per video, extracted at one frame per second. 

\begin{figure*}
    \centering
    \includegraphics[width=0.95\textwidth]{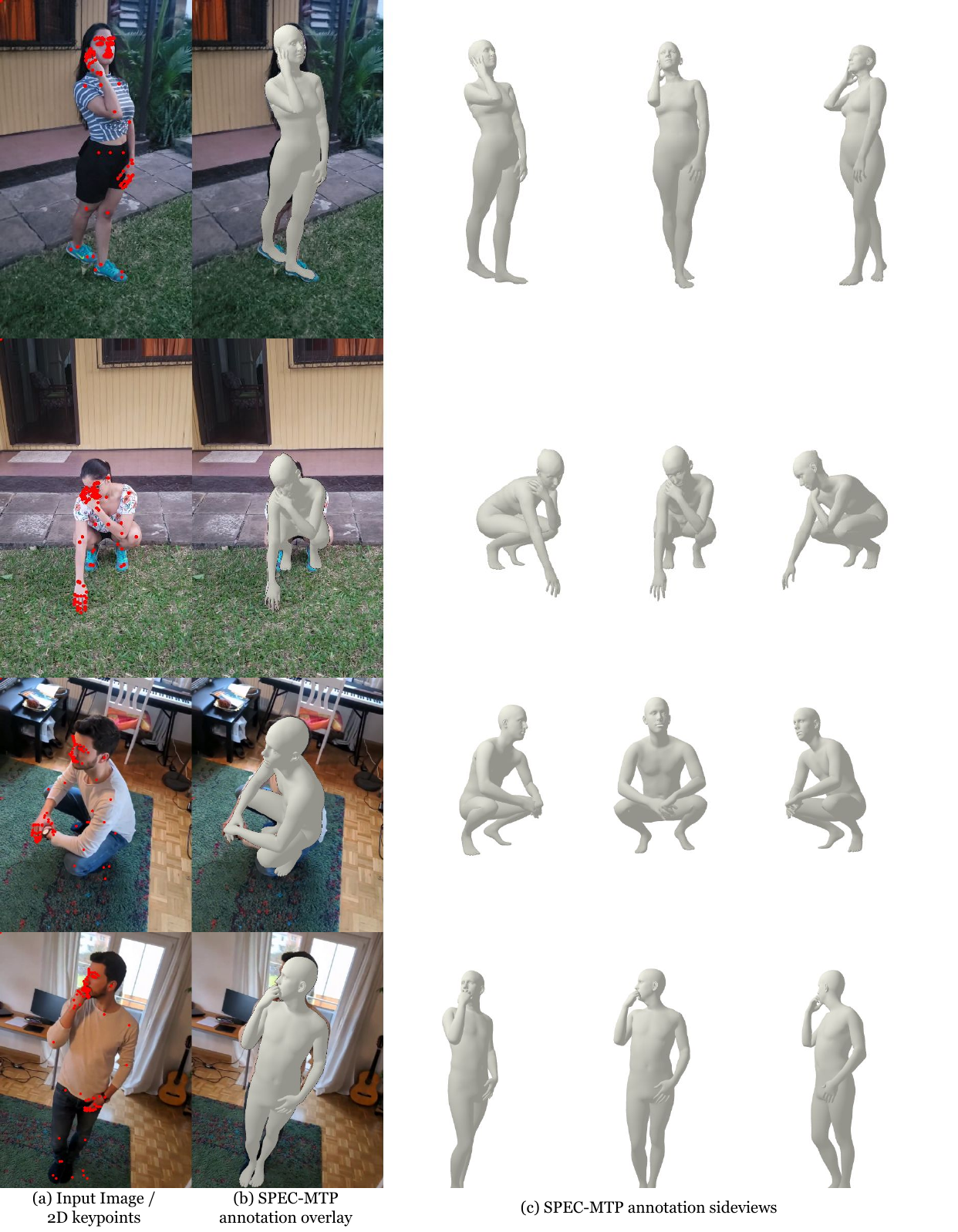}
    \caption{\textbf{\mtpcam benchmark samples.}}
    \label{fig:mtp-cam}
\end{figure*}

In the second and the third stage, we freeze body shape $\beta$ and refine the pose $\theta$ and camera parameters by minimizing: 
\begin{align}
\begin{split}
E(\theta, \pitch, \roll, \yaw, \camtransl) = & \lambda_{m_h}E_{m_h} + \lambda_{\tilde{\theta}} E_{\tilde{\theta}} + \lambda_{\tilde{C}} E_{\tilde{C}} + \\
                                        & \lambda_{S} E_{S} + \sum_{i=1}^{F} E_{J_i} \text{.}
\end{split}
\end{align}
$E_{m_h}$, $E_{\tilde{\theta}}$, $E_{\tilde{C}}$, $E_{S}$ denote the hand and presented pose priors, the presented contact loss and the general contact loss as defined in \cite{Mueller:CVPR:21}, respectively.
Fig.~\ref{fig:mtp-cam} shows several examples of \mtpcam frames and the computed SMPL-X fit.
\mtpcam is used only for evaluation.

\subsection{\agoracam Dataset}
\begin{figure*}
    \centering
    \includegraphics[width=0.85\textwidth]{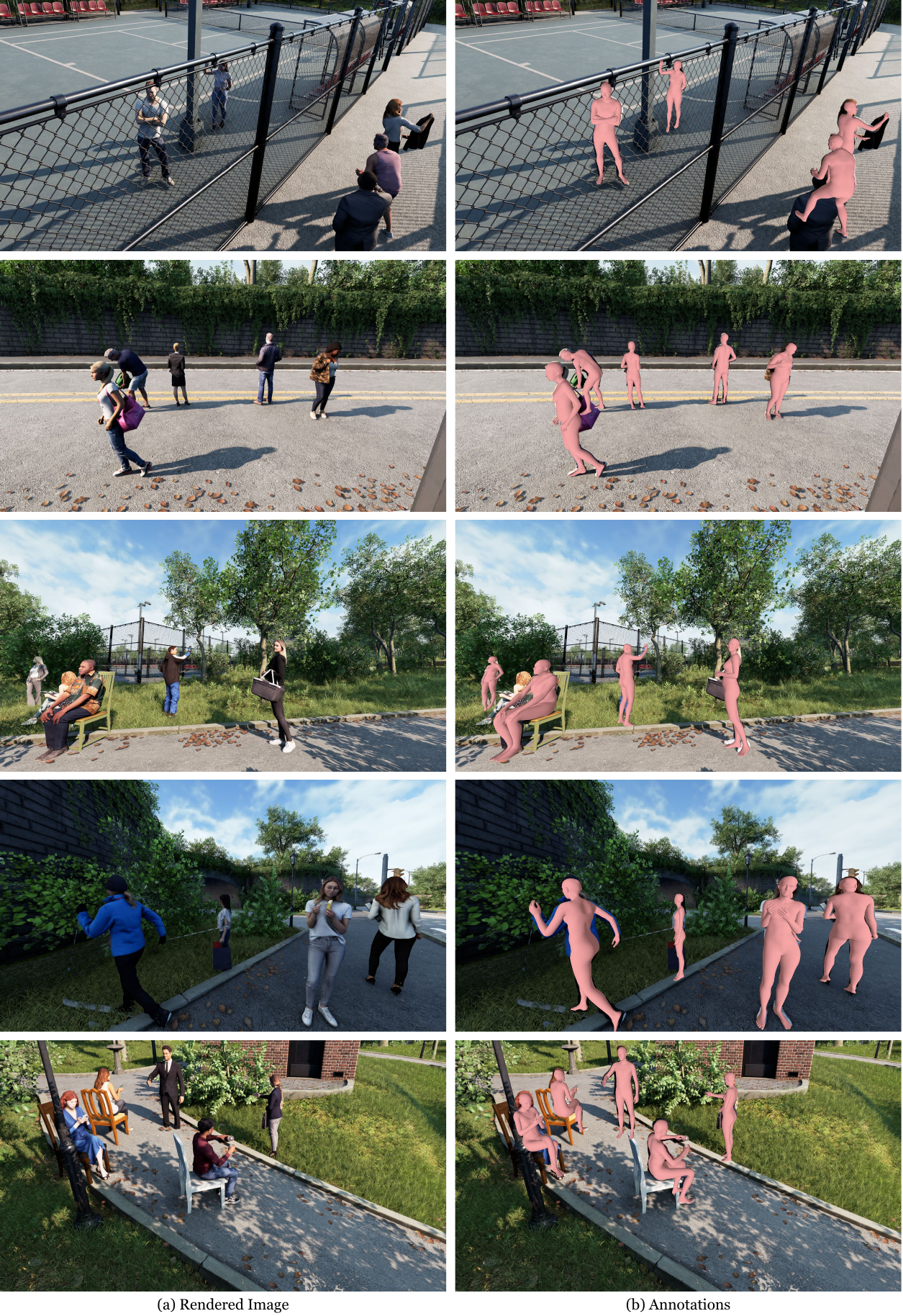}
    \caption{\textbf{\agoracam dataset samples.}}
    \label{fig:agora-cam-samples}
\end{figure*}

We obtain the 3D scans and SMPL-X fits to those scans from the AGORA dataset \cite{patel2021agora}. 
This includes many high-quality 3D scans of clothed people with accurate SMPL-X ground truth shape and pose. 
We convert the SMPL-X model to the SMPL format.
We then put these scans in 3D scenes and use Unreal engine \cite{unrealengine} to generate photorealistic images with diverse fields of view (fov) and camera rotations.
Fig.~\ref{fig:agora-cam-samples} shows several examples, with SMPL fits overlaid on the images.
One can observe some perspective distortion at the image boundary in the 3rd and 4th rows, indicating large fov (small focal length); the first and the last row show examples with high camera pitch angles.

\subsection{Pano360 Dataset}

\begin{figure*}
    \centering
    \includegraphics[width=0.9\textwidth]{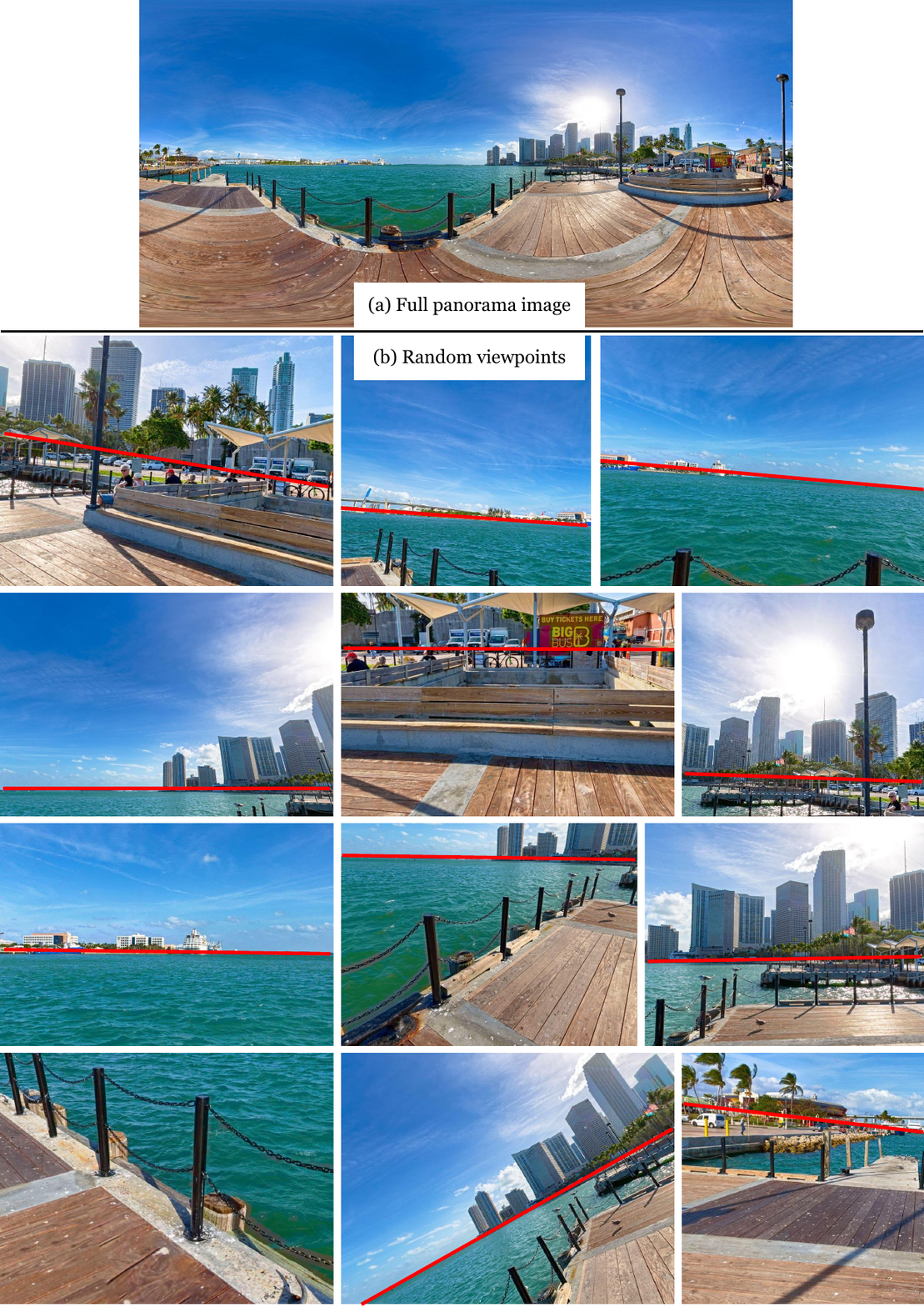}
    \caption{\textbf{Pano360 dataset.} Random viewpoints (b) from a single equirectangular panorama image (a). Horizon annotations are shown in red line.}
    \label{fig:pano360}
\end{figure*}

To generate training dataset from equirectangular panorama images, we follow the strategy of Zhu \etal~\cite{zhu2020single}. We crop images from panorama images with random viewpoints and focal lengths. Fig.~\ref{fig:pano360} shows a sample panorama image along with the cropped images. 

\section{Experiments}
%%%%%%%%%%%%%%%%%%%%%%%%%% TABLE 2/3/4 WITH MPJPE %%%%%%%%%%%%%%%%%%%%%%%%%%

\subsection{Training Datasets}

In addition to \agoracam dataset which is described in main text, we use datasets explained below for training.

\noindent\textbf{MPI-INF-3DHP}~\cite{mpiiinf3dhp_mono-2017} is a multi-view indoor 3D human pose estimation dataset. 3D annotations are captured via a commercial markerless mocap software, therefore it is less accurate than some of the 3D datasets \eg \hthreesixm~\cite{ionescu_h36m}. We use all of the training subjects S1 to S8 which makes 90K images in total.

\noindent\textbf{Human3.6M}~\cite{ionescu_h36m} is an indoor, multi-view 3D human pose estimation dataset. Following previous methods, for training, we use 5 subjects (S1, S5, S6, S7, S8) which means 292K images.

\noindent\textbf{COCO}~\cite{coco} dataset is a 2D keypoint dataset. In addition to 2D keypoint annotations, we utilize SMPLift-X-cam and \camcalib method to obtain SMPL and camera parameters annotations. We initialize the SMPLify-X-cam with SMPL fits provided by EFT~\cite{joo2020eft} method.

\paragraph{Training Dataset Ratios.}
To obtain the final best performing model, we follow EFT~\cite{joo2020eft} and SPIN~\cite{SPIN:ICCV:2019} which use fixed data sampling ratios for each batch. We first train SPEC with 50\% \agoracam, 50\% COCO for 175K steps. Then, we continue training 
with 20\% Human3.6M, 20\% MPI-INF-3DHP, with 50\% \agoracam, and 50\% COCO for around 50K steps until convergence.

\subsection{CamCalib Qualitative Results}
\begin{figure*}
    \centering
    \includegraphics[width=0.92\textwidth]{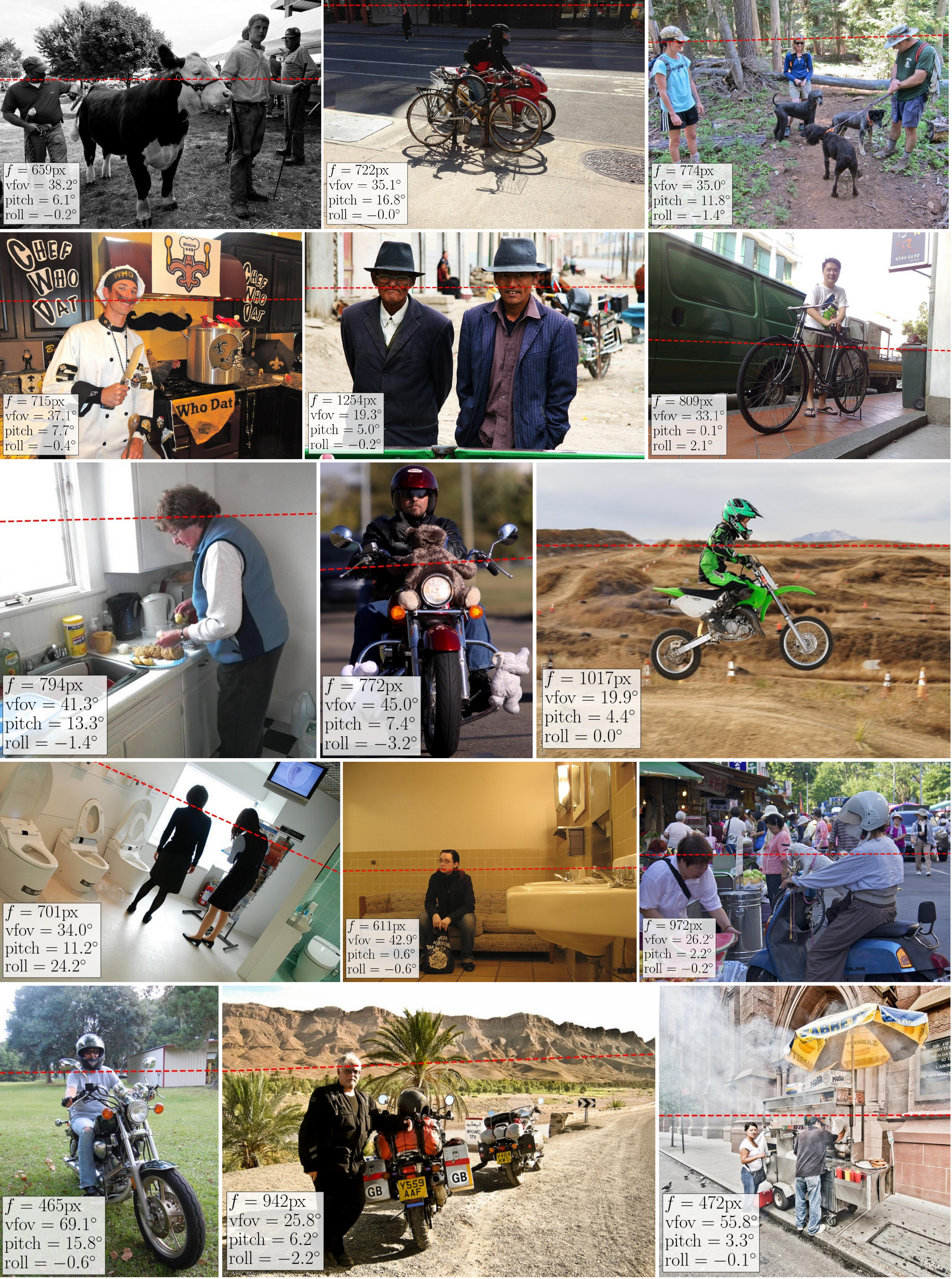}
    \caption{\textbf{\camcalib qualitative results.}}
    \label{fig:camcalib_qual_good}
\end{figure*}

\begin{figure*}
    \centering
    \includegraphics[width=0.992\textwidth]{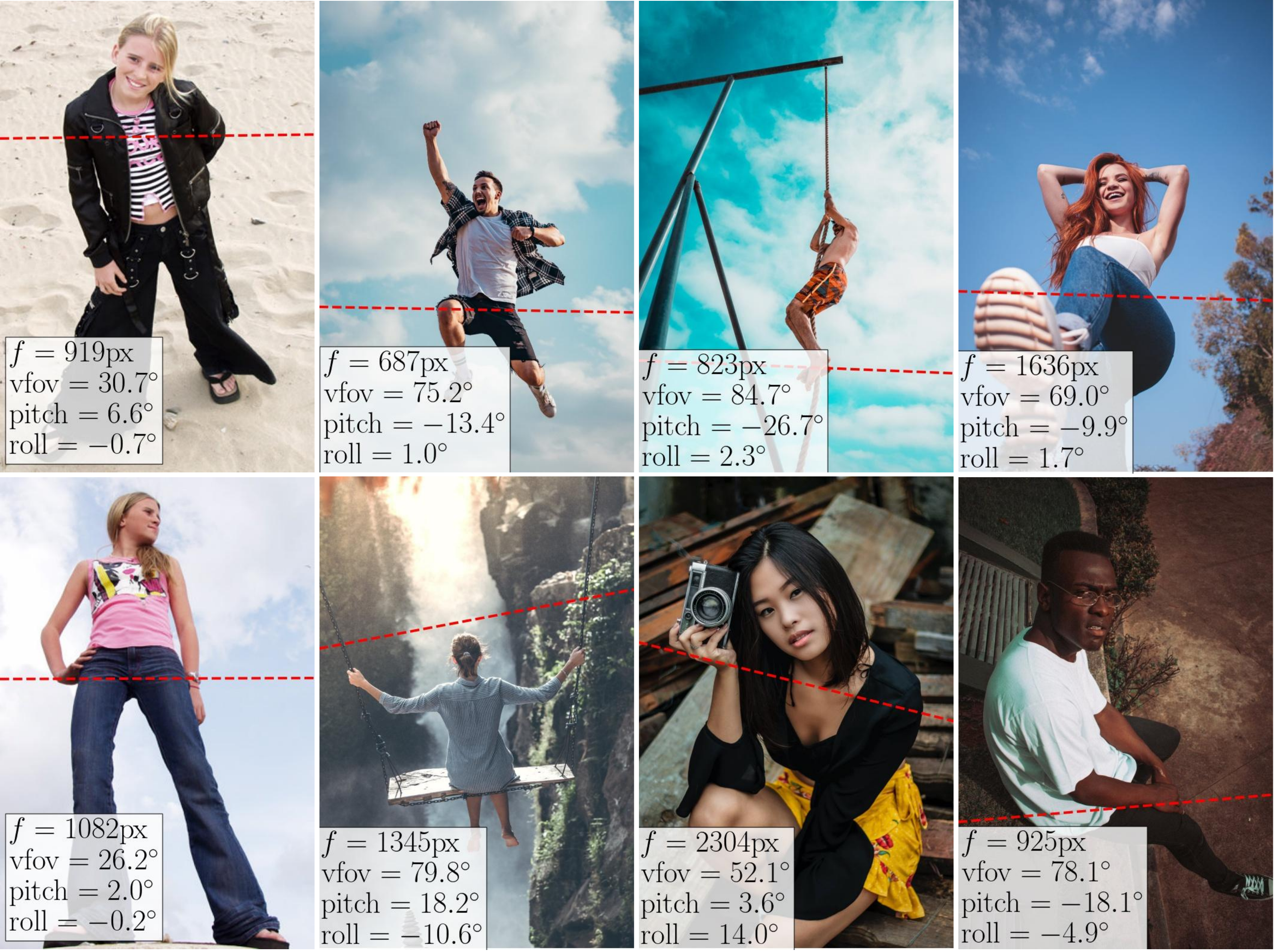}
    \caption{\textbf{\camcalib failures.}}
    \label{fig:camcalib_qual_bad}
\end{figure*}

In Fig.~\ref{fig:camcalib_qual_good}, we show the qualitative results of \camcalib. 
We follow \cite{zhu2020single} to visualize the estimated camera rotation by drawing the estimated horizons (red dashed lines). 
If the camera is pointing down (pitch angle $\pitch > 0$), the horizon should appear in the upper half of the image. 
Tilting to the left or right indicates the camera roll.
\camcalib estimates reasonable camera parameters for most examples.
We also show the failure cases in Fig.~\ref{fig:camcalib_qual_bad}.
We observe that they are all portrait images in which background contain little information for estimating camera parameters.
We remark that despite no rich information in the background, human bodies still provide useful cues for the calibration purpose and we leave this to the future work.

\subsection{\methodname MPJPE/PVE Results and Discussions}
Table \ref{tab:sota_mtp} to \ref{tab:sota_3dpw} summarize the performance of \methodname in comparison to SOTA methods on three datasets: \mtpcam, \agoracam, and 3DPW.
In addition to the three metrics, W-MPJPE, PA-MPJPE and W-PVE that are already reported in the main paper, we also include MPJPE and PVE here. 
The two versions of W-MPJPE and W-PVE are defined in Sec.~4.2 in the main text. 

First, we observe that \methodname yields better ``pure body pose'' according to the improved PA-MPJPE.
Moving onward, a 3DHPS method should learn not only to reconstruct body poses and shapes but also to place and orient them properly in the space.
To this end, MPJPE and PVE are often considered stricter than the Procrustes-aligned counterparts as they measure additionally discrepancies in rotation. 
\methodname also outperforms SOTA methods \cite{SPIN:ICCV:2019,kolotouros2019cmr,Moon_2020_ECCV_I2L-MeshNet,Rockwell2020} in MPJPE/PVE, but yield on-par or slightly worse results than HMR$^*$, which is an \iwcam baseline trained under the identical setting as \methodname.

Note that MPJPE and PVE are typically computed in the camera space. 
One needs to transform the ground truth bodies using the camera extrinsics provided by the datasets, and thus the error encodes dataset-specific camera information.
However, existing benchmarks, \eg, MPI-INF-3DHP, Human3.6M and 3DPW are often captured with little variation in camera parameters.
As a results, MPJPE and PVE cannot clearly reflect the performance of a \iwcam method for in-the-wild scenarios where camera types and viewpoints are diverse and unknown.
We advocate W-MPJPE and W-PVE because they also measure discrepancies in rotation, but unlike MPJPE/PVE, they are computed in world coordinates, assuming no access to camera information. 
In W-MPJPE and W-PVE, \methodname again outperforms SOTA methods \cite{SPIN:ICCV:2019,kolotouros2019cmr,Moon_2020_ECCV_I2L-MeshNet,Rockwell2020} and yield consistently better results than HMR$^*$ on datasets captured with diverse camera parameters -- \mtpcam and \agoracam. Even on 3DPW which is captured with single focal length value, \methodname attains improved or on-par results than HMR$^*$.

\input{tables/supmat_mtp_j24}
\input{tables/supmat_agora_j24}
\input{tables/supmat_3dpw}

%%%%%%%%%%%%%%%%%%%%%%%%%% SMPLIFY-X-CAM RESULTS %%%%%%%%%%%%%%%%%%%%%%%%%%%
\subsection{\smplify Results}
\label{sec:smplifycam_evaluation}
Table \ref{tab:smplify_agora} and \ref{tab:smplify_3dpw} summarize the performance of \smplify in comparison to the baseline SMPLify-X method. 
The same 2D keypoint detections from \cite{mmpose2020} are used for all the reported methods. 
We compare the default SMPLify-X with $f=5000$ \cite{SMPL-X:2019}, the setting considered by Kissos \etal~\cite{kissosECCVW2020} where $f=2200$, and the setting which uses \camcalib estimated $K$ and $\camrot$. 
We observe a consistent improvement in \wmpjpe and \wpve when $\camrot$ is used. 
This is due to correct global orientation reconstruction w.r.t.~world coordinates. 
On \agoracam, using $K$ improves the \pampjpe due to more accurate projective geometry. 
3DPW is captured with a single camera where $f=1962$ so it is not a good dataset with which to evaluate the effect of focal length.
Even in this case, \smplify improves the results over the default setting and the \pampjpe result is on par with results using $f=2200$. 
The $f=2200$ approximation is already close to the single focal length used in 3DPW dataset $f=1962$, consequently it does well.
As a reference, the average \camcalib focal length error is 360 and 246 pixels on 3DPW and \agoracam datasets, respectively.

\begin{figure}
    \centering
    \includegraphics[width=0.45\textwidth]{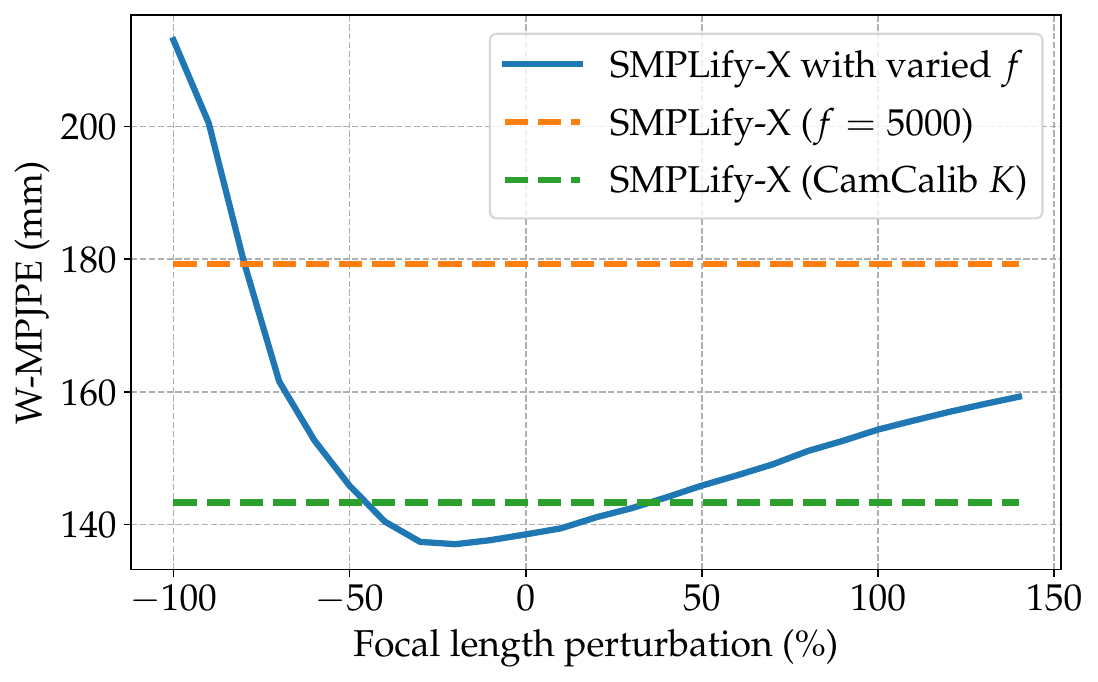}
    \caption{Sensitivity of SMPLify-X to focal length perturbation on \agoracam dataset. Using \camcalib estimated $f$ yields better results. Body reconstruction accuracy is less sensitive to larger focal lengths. Therefore, we propose \softbiasedltwo loss.}
    \label{fig:f_sensitivity}
\end{figure}

To further analyze the impact of focal length on body reconstructions, we run SMPLify-X on \agoracam with focal lengths perturbed from the real ground truth values
and plot the W-MPJPE trend in Fig.~\ref{fig:f_sensitivity} (blue curve).
We see that the quality of HPS are sensitive to underestimated focal lengths and less sensitive to overestimation, as also reported in \cite{kissosECCVW2020,yu2020pcls}.
In addition, we visualize default SMPLify-X ($f=5000$) and SMPLify-X (\camcalib $K$) according to the corresponding W-MPJPE in Table \ref{tab:smplify_agora}.
One can see that in average, with the focal lengths from \camcalib, body reconstructions are closer to the low-error basin of small perturbations, \ie~closer to the true focal lengths. 
Note that, the averaged focal length in \agoracam is 840 pixels, so the averaged error of 246 pixels (29\%) confirms again that \smplify is relatively robust when the estimated focal length ranges between 0.7 and 1.3 times the true one.
\input{tables/supmat_smplify_agora_j24}

\input{tables/supmat_smplify_3dpw}

\subsection{Ablation study on \methodname with 3DPW}
The Table 5 in the main text provides an ablation study on \methodname using the \agoracam dataset, 
which dissects the improvement over the baseline HMR$^*$ in various aspects: using the original image center as the principal point, using the \camcalib estimated focal length, using the estimated rotations, and lastly conditioning the network with the estimated cameras (\methodname).
We repeat this here on the common 3DPW benchmark as in Table \ref{tab:hmr_3dpw}.
Despite that it is not a suitable dataset to analyze the impact of each camera parameters, we still observe that appending the estimated cameras to the image feature leads to improvement in five metrics (c.f. the last two rows), 
so does using the estimated focal length (HMR$^{*}$ + $c$ + $f$ vs.  HMR$^{*}$ + $c$).

% \textcolor{red}{Explain if there are differences.  What should we pay attention to?}

\begin{table}
    \centering
    \resizebox{0.47\textwidth}{!}{
        \begin{tabular}{l|r|r|r}
        \toprule
        \textbf{Methods} & \textbf{\wmpjpe} & \textbf{\pampjpe} & \textbf{\wpve} \\ 
        \midrule
        HMR$^{*}$                                 & 112.7 / \textbf{97.2} & 62.3 & 133.1 / 115.6 \\
        HMR$^{*}$ + $c$                           & 115.4 / 97.3 & 62.2 & 135.9 / 116.9 \\
        HMR$^{*}$ + $c$ + $f$                     & 112.1 / 96.1 & 61.4 & 134.1 / \textbf{113.6} \\
        \midrule
        HMR$^{*}$ + $c$ + $f$ + $\camrot$         & 102.7 / 102.7 &  60.2 & 124.0 / 124.0 \\
        SPEC & \textbf{98.1} / 98.1 & \textbf{59.9} & \textbf{119.3} / 119.3 \\
        % SPEC - cam cond. & & & \\ 
        \bottomrule
        % HMR + cam center + gt fl + gt rot & & & \\ \hline
        \end{tabular} 
    }
    \caption{Ablation experiments on \methodname with 3DPW validation set. $c$: using the image center as camera center; $f$ and $\camrot$: using \camcalib estimated focal length and camera rotation, respectively. All numbers are in \emph{mm}.}
    \label{tab:hmr_3dpw}
\end{table}

\begin{table}
    \centering
    \resizebox{0.3\textwidth}{!}{
        \begin{tabular}{l|r|r}
        \toprule
        \textbf{Methods} & \textbf{\mpjpe} & \textbf{\pampjpe} \\ 
        \midrule
        Want~\etal~\cite{wang2020viewpoint}& 89.7 & 65.2 \\
        SPEC w. 3DPW & 96.4 & 52.7 \\
        \bottomrule
        \end{tabular} 
    }
    \caption{Comparison to Wang~\etal~\cite{wang2020viewpoint}. Here both methods are trained with 3DPW training set for a fair comparison.}
    \label{tab:comparison_wang_etal}
\end{table}

\subsection{Comparison to Wang \etal~\cite{wang2020viewpoint}}
Wang~\etal~\cite{wang2020viewpoint} train their methods on 3DPW training set. We also train SPEC on 3DPW to make a comparison to their method. Results are denoted in Table~\ref{tab:comparison_wang_etal}. SPEC outperforms Wang~\etal~\cite{wang2020viewpoint} in terms of PA-MPJPE metric, but performs poorly in MPJPE. This is due to the use of estimated camera parameters in SPEC's evaluation. We argue that SPEC would perform better with W-MPJPE metric, however a comparison is not possible since the code of Wang~\etal~\cite{wang2020viewpoint} is not available.

\subsection{Qualitative results of \methodname}
\begin{figure*}
    \centering
    \includegraphics[width=\textwidth]{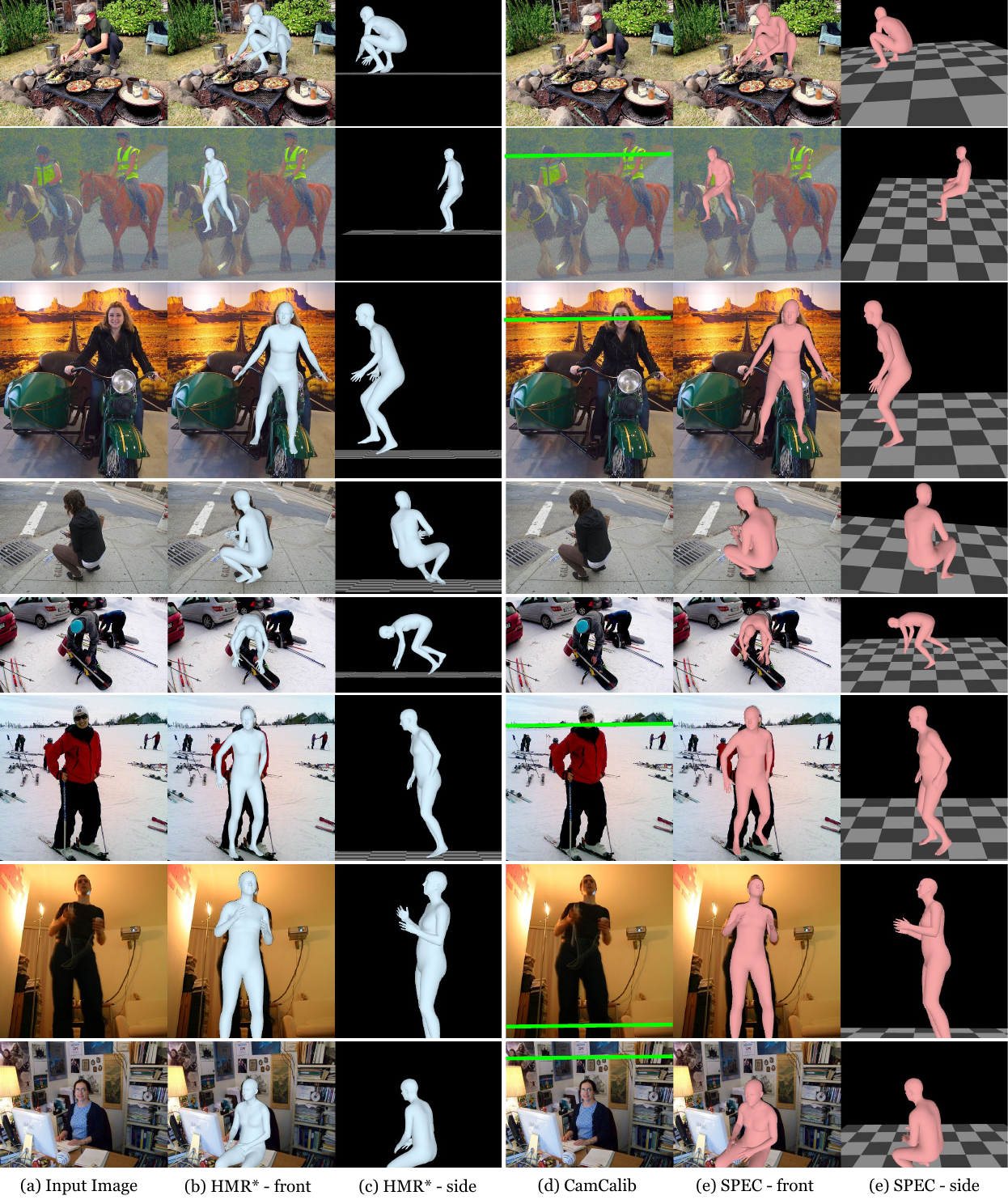}
    \caption{\textbf{SPEC qualitative results.}}
    \label{fig:spec_qualitative_good}
\end{figure*}

In Fig.~\ref{fig:spec_qualitative_good}, we show several qualitative results from \methodname. 
One can observe that \methodname yields on-par or more physically plausible reconstructed bodies than the baseline that is trained with the identical setting.
For more clear illustration, please see the 360$^\circ$ visualizations in \supmat~video.

\begin{figure*}
    \centering
    \includegraphics[width=\textwidth]{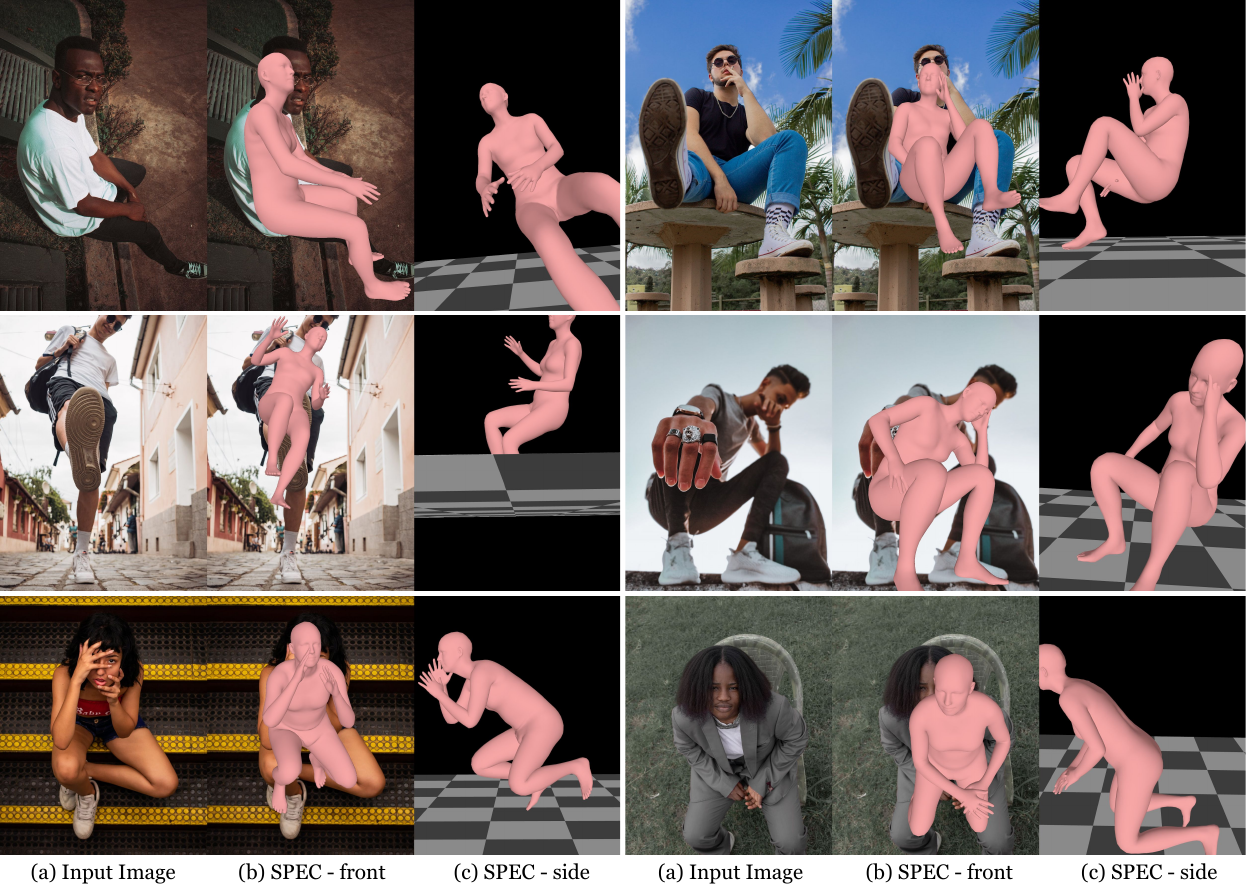}
    \caption{\textbf{SPEC failures.}}
    \label{fig:spec_qualitative_bad}
\end{figure*}

The failure cases of \methodname are shown in Fig.~\ref{fig:spec_qualitative_bad}. 
We observe that some examples share similar traits as those in Fig.~\ref{fig:camcalib_qual_good}: portrait images with limited background information. 
The error of \methodname can be partially attributed to the error from \camcalib.
Other scenarios include rarely-seen viewpoints or poses that are not observed in the training data.
%The errors of \camcalib could potentially be carried over to \methodname. 

%% file: tables/supmat_mtp_j24.tex
\begin{table}
    \centering
    \resizebox{0.47\textwidth}{!}{
        \begin{tabular}{l|r|r|r|r|r}
        \toprule
        \textbf{Methods} & \textbf{\mpjpe} & \textbf{\wmpjpe} & \textbf{\pampjpe} & \textbf{\wpve} & \textbf{\pve}\\ 
        \midrule 
        GraphCMR~\cite{kolotouros2019cmr} & 150.9 & 175.1 / 166.1 & 94.3 & 205.5 / 197.3 & 179.9 \\
        SPIN~\cite{SPIN:ICCV:2019} & 129.4 & 143.8 / 143.6 & 79.1 & 165.2 / 165.3 & 148.2 \\
        % Pose2Mesh~\cite{Choi_2020_ECCV_Pose2Mesh} & & & & & \\
        PartialHumans~\cite{Rockwell2020} & 150.1 & 158.9 / 157.6 & 98.7 & 190.1 / 188.9 & 177.8 \\
        I2L-MeshNet$^{\dagger}$~\cite{Moon_2020_ECCV_I2L-MeshNet} & 155.5 & 167.2 / 167.0 & 99.2 & 199.0 / 198.1 & 184.4 \\
        % CenterHMR~\cite{CenterHMR} & 185.5 & 70.3 & 197.4 \\
        % CRMH~\cite{jiang2020mpshape} & & & & & \\ 
        \midrule
        HMR$^{*}$~\cite{kanazawa_hmr} &  \textbf{109.0} & 142.5 / 128.8 & \textbf{71.8} & 164.6 / 150.7 & \textbf{127.6} \\
        \methodname & 116.1 & \textbf{124.3} / \textbf{124.3} & \textbf{71.8} & \textbf{147.1} / \textbf{147.1} & 136.4 \\
        \bottomrule
        \end{tabular} 
    }
    \caption{Results of SOTA methods on \mtpcam dataset. 
    We use the implementations provided by the authors to obtain results. 
    HMR$^{*}$ means that we train HMR using the same data as \methodname for fair comparison. $^{\dagger}$means we use the SMPL output of this method instead of the non-parametric mesh to be able to report \wpve. All numbers are in \emph{mm}. % \mkocabas{Evaluation protocol is updated to use 24 joints.}
    %We use the implementations provided by authors to obtain results. HMR$^{*}$ denotes that we train HMR using the same data as \methodname to make a fair comparsion. $^{\dagger}$ we use the SMPL output of this method instead of non-parametric mesh to be able to report \wpve.
    }
    \vspace{-1ex}
    \label{tab:sota_mtp}
\end{table}

%% file: tables/supmat_agora_j24.tex
\begin{table}
    \centering
    \resizebox{0.47\textwidth}{!}{
        \begin{tabular}{l|r|r|r|r|r}
        \toprule
        \textbf{Methods} & \textbf{\mpjpe} & \textbf{\wmpjpe} & \textbf{\pampjpe} & \textbf{\wpve} & \textbf{\pve}\\ 
        \midrule 
        GraphCMR~\cite{kolotouros2019cmr} & 179.8 & 181.7 / 181.5 & 86.6 & 219.8 / 218.3 & 216.8\\
        SPIN~\cite{SPIN:ICCV:2019} & 159.6 & 165.8 / 161.4 & 79.5 & 194.1 / 188.0 & 186.3 \\
        % Pose2Mesh~\cite{Choi_2020_ECCV_Pose2Mesh} & & & & & \\
        PartialHumans~\cite{Rockwell2020} & 172.6 & 169.3 / 174.1 & 88.2 & 207.6 / 210.4 & 209.0 \\
        I2L-MeshNet$^{\dagger}$~\cite{Moon_2020_ECCV_I2L-MeshNet} & 161.7 & 169.8 / 163.3 & 82.0 & 203.2 / 195.9 & 194.5 \\
        % CenterHMR~\cite{CenterHMR} & 185.5 & 70.3 & 197.4 \\
        % CRMH~\cite{jiang2020mpshape} & & & & & \\ 
        \midrule
        HMR$^{*}$~\cite{kanazawa_hmr} & 92.8 & 128.7 / 96.4 & 55.9 & 144.2 / 111.8 & 108.1\\
        \methodname &  \textbf{74.9} & \textbf{74.9} / \textbf{74.9} & \textbf{54.5} & \textbf{90.5} / \textbf{90.5} & \textbf{90.5} \\
        \bottomrule
        \end{tabular} 
    }
    \caption{Results of SOTA methods on \agoracam. See Table~\ref{tab:sota_mtp} caption. % \mkocabas{Evaluation protocol is updated to use 24 joints.}
    }
    \vspace{-1ex}
    \label{tab:sota_agora}
\end{table}

%% file: tables/supmat_3dpw.tex
\begin{table}[]
    \centering
    \resizebox{0.49\textwidth}{!}{
        \begin{tabular}{l|r|r|r|r|r}
        \toprule
        \textbf{Methods} & \textbf{\mpjpe} & \textbf{\wmpjpe} & \textbf{\pampjpe} & \textbf{\wpve} & \textbf{\pve} \\ 
        \midrule 
        % GraphCMR~\cite{kolotouros2019cmr} & 198.324 & 197.182 & 79.779 & 215.141 & 227.888 \\
        GraphCMR~\cite{kolotouros2019cmr} & 121.2 & 137.8 / 129.4 & 69.1 & 158.4 / 152.1 & 139.3 \\
        SPIN~\cite{SPIN:ICCV:2019} & 96.9 & 122.2 / 116.6 & 59.0 & 140.9 / 135.8 & 129.8 \\
        Partial Humans~\cite{Rockwell2020} & 126.5 & 139.4 / 132.9 & 76.9 & 160.1 / 152.7 & 144.5 \\
        I2L-MeshNet$^{\dagger}$~\cite{Moon_2020_ECCV_I2L-MeshNet} & 100.0 & 133.3 / 119.6 & 60.0 & 154.5 / 141.2 & 129.5 \\
        % CenterHMR~\cite{CenterHMR} & 130.3 & 60.8 & 149.3 \\
        \midrule
        HMR$^{*}$~\cite{kanazawa_hmr} & \textbf{92.5} & 119.2 / \textbf{104.0} & 53.7 & 136.2 / \textbf{120.6} & \textbf{109.5} \\
        \methodname  & 96.5 & \textbf{106.4} / 106.4 & \textbf{53.2} & \textbf{127.4} / 127.4 & 118.5 \\ 
        \bottomrule
        \end{tabular} 
    }
    \caption{Results of SOTA methods on 3DPW test set. See Table~\ref{tab:sota_mtp} caption.
    %We use the implementations provided by authors to obtain results. HMR$^{*}$ denotes that we train HMR using the same data as \methodname to make a fair comparsion. $^{\dagger}$ we use the SMPL output of this method instead of non-parametric mesh to be able to report \wpve.
    }
    \label{tab:sota_3dpw}
\end{table}

%% file: tables/supmat_smplify_agora_j24.tex
\begin{table}[]
    \centering
    \resizebox{0.48\textwidth}{!}{
        \begin{tabular}{l|r|r|r}
        \toprule
        \textbf{Methods} & \textbf{\wmpjpe} & \textbf{\pampjpe} & \textbf{\wpve} \\ \midrule
        SMPLify-X (ground-truth $f$) &  131.9 / 114.6 & 73.3 & 151.5 / 133.4\\ 
        \midrule
        SMPLify-X \cite{SMPL-X:2019} ($f=5000$) & 168.9 / 149.5 & 77.1 & 191.5 / 172.8 \\ 
        SMPLify-X \cite{kissosECCVW2020} ($f=2200$) &  155.6 / 133.5 & 75.6 & 176.9 / 155.3\\ 
        \midrule
        SMPLify-X (\camcalib $K$) & 136.5 / 116.4 & \textbf{73.0} & 156.3 / 135.8 \\
        \smplify (\camcalib $K$ + $\camrot$) &\textbf{ 115.2} / \textbf{115.2} & 73.5 & \textbf{135.0} / \textbf{135.0} \\ 
        \bottomrule
        \end{tabular} 
    }
    \caption{{\bf HPS optimization with an estimated camera.}  \smplify on the \agoracam dataset.} % \mkocabas{Evaluation protocol is updated to use 24 joints.}}
    \label{tab:smplify_agora}
\end{table}

%% file: tables/supmat_smplify_3dpw.tex
\begin{table}[]
    \centering
    \resizebox{0.48\textwidth}{!}{
        \begin{tabular}{l|r|r|r}
            \toprule
            \textbf{Methods} & \textbf{\wmpjpe} & \textbf{\pampjpe} & \textbf{\wpve} \\ \midrule 
            % PARE & 111.01 & 53.94 & 130.15 \\ \hline
            % SMPLify-x GT FL & 124.898 & 55.389 & 146.93 \\ \midrule
            
            % SMPLify-x Default & 123.843 & 57.668 & 144.833 \\
            % SMPLify-x Amazon FL & 124.809 & 55.485 & 146.651 \\ \midrule
            
            % SMPLify-x pred fl & 125.498 & 55.643 & 147.37 \\
            % SMPLify-x pred fl + rot & \textbf{95.538} & 55.801 & \textbf{120.377} \\
            
            SMPLify-X (ground-truth $f$) & 124.9 / 95.0 & 55.4 & 146.9 / 120.0 \\ 
            \midrule  
            SMPLify-X \cite{SMPL-X:2019} ($f=5000$) & 123.8 / 94.7 & 57.7 & 144.8 / 119.8 \\
            SMPLify-X \cite{kissosECCVW2020} ($f=2200$) & 124.8 / \textbf{94.4} & \textbf{55.5} & 146.7 / \textbf{119.6} \\ 
            \midrule 
            SMPLify-X (\camcalib $K$) & 125.5 / 95.4 & 55.7 & 147.4 / 120.3 \\
            \smplify (\camcalib $K$ + $\camrot$) & \textbf{95.5} / 95.5 & 55.8 & \textbf{120.4} / 120.4 \\ 
            \bottomrule
        \end{tabular} 
    }
    \caption{{\bf HPS optimization with an estimated camera.}  \smplify on the 3DPW validation set.}
    \label{tab:smplify_3dpw}
\end{table}